\documentclass[runningheads]{llncs}

\usepackage{eccv}

\usepackage{eccvabbrv}

\usepackage{graphicx}
\usepackage{booktabs}
\usepackage{multirow}   % For multi-row cells
\usepackage{xcolor}     % For row colors and text colors
\usepackage{colortbl}   % For coloring table rows
\usepackage{makecell}
\usepackage{graphicx}

\definecolor{rowgreen}{RGB}{228, 237, 236}  % 淡颜色绿
\definecolor{rowgray}{RGB}{235, 235, 235}   % 闭源/专有模型 (灰)

\usepackage[accsupp]{axessibility}  % Improves PDF readability for those with disabilities.

\usepackage[pagebackref,breaklinks,colorlinks,citecolor=eccvblue]{hyperref}
\usepackage{orcidlink}

\usepackage[skins]{tcolorbox}
\usepackage{tabularx} % 自动调整列宽的宏包

\newtcolorbox{promptbox}[1]{
    enhanced,
    colback=rowgreen,       % 框内背景色：极浅的蓝紫色
    colframe=black,             % 外边框颜色：黑色
    boxrule=1.5pt,              % 外边框粗细
    arc=6pt,                    % 边框圆角大小
    title=#1,                   % 传入标题内容
    fonttitle=\bfseries\large,  % 标题字体：加粗、放大
    coltitle=white,             % 标题文字颜色：白色
    attach boxed title to top left={xshift=15pt, yshift=-\tcboxedtitleheight/2},
    boxed title style={
        colback=black,          % 标题框背景色：黑色
        colframe=black,         % 标题框边框颜色
        arc=3pt,                % 标题框圆角
        boxrule=0pt,
        left=8pt, right=8pt, top=4pt, bottom=4pt % 标题框内的内边距
    },
    left=15pt, right=15pt, top=15pt, bottom=15pt
}

\begin{document}

% \title{: Structured Chain-of-thought Utilizing Process-Supervised RL Training for Spatial Reasoning} 
\title{SCOUT: Unlocking Enhanced Spatial Reasoning via Structured Chain-of-Thought and Multi-Objective Process Reward} 


\titlerunning{SCOUT}

\author{Zile Zhou \and
Huining Yuan \and
Weichen Zhang \and Xinlei Chen \and Xiao-ping Zhang}

\authorrunning{Z. Zhou et al.}


\institute{Shenzhen International Graduate School, Tsinghua University, Shenzhen, China}

% \email{jilokchow@gmail.com, \{yuanhu\}@sz.tsinghua.edu.cn, }

% Structured Chain-Of-Thought Utilizing Process-Supervised RL Training

\maketitle

\begin{abstract}

Existing Vision-Language Models (VLMs) exhibits a critical bottleneck in robust spatial reasoning.
Recent reinforcement learning (RL) methods aim to close this gap with verifiable outcomes, yet they suffer from poor credit assignment across intermediate reasoning steps.
Concurrently, structured reasoning approaches overlook the critical depth perception necessary for comprehensive 3D understanding.
To address these challenges, we propose \textbf{SCOUT} (\underline{S}tructured \underline{C}hain-\underline{O}f-Thought \underline{U}tilizing Process-Supervised RL \underline{T}raining). Specifically, we design a structured Chain-of-Thought (CoT) framework that explicitly models 3D environmental perception to ensure robust spatial understanding and reasoning. Furthermore, we introduce a novel RL algorithm featuring multi-objective process rewards and a tailored advantage estimation method, facilitating fine-grained credit assignment across distinct segments of the reasoning trajectory. To support our framework, we develop SCOUT-24k, a structured spatial reasoning CoT dataset synthesized through a customized pipeline. Extensive evaluations demonstrate that SCOUT-3B improves upon baseline models by 16.85\% and 6.3\% on general spatial benchmarks and complex spatial reasoning tasks respectively. 
Notably, our larger SCOUT-7B even outperforms GPT-4o by a margin of 4.28\%.
Moreover, despite being trained exclusively on single image, SCOUT-7B exhibits robust out-of-domain generalization to multi-image and video scenarios. These empirical results render SCOUT as a critical step towards next generation of spatially-aware VLMs.


\end{abstract}

\begin{figure*}[tb]
  \centering
  \includegraphics[width=\textwidth]{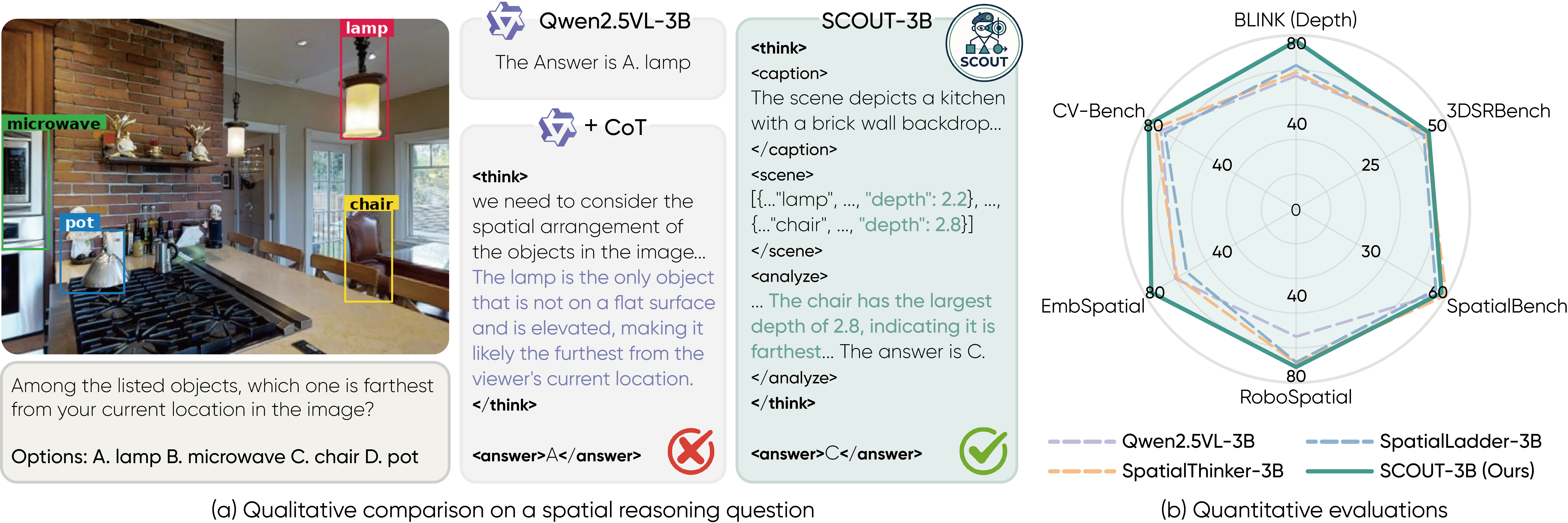}
  \caption{ 
We present \textbf{SCOUT}, a novel spatial reasoning framework for VLMs, equipped with a structured CoT optimized via process-supervised RL. \textbf{(a) Qualitative example:} the baseline produces an incorrect answer due to a lack of spatial awareness, whereas our model generates accurate prediction by explicitly analyzing object depth and scene information. \textbf{(b) Quantitative evaluations} on multiple spatial reasoning benchmarks demonstrate the superiority of SCOUT-3B over Qwen2.5-VL-3B and existing spatial-specialized methods.
  }
  \label{fig:teaser}
\end{figure*}

\section{Introduction}

In recent years, Vision-Language Models (VLMs) have achieved remarkable success in fundamental visual tasks~\cite{bai2025qwen2, hurst2024gpt, comanici2025gemini}. 
Despite this progress, in many critical downstream applications that require advanced spatial reasoning, such as robot navigation~\cite{liu2023aerialvln, yokoyama2024vlfm}, robot manipulation~\cite{driess2022learning, huang2024rekep, zitkovich2023rt}, autonomous driving~\cite{tian2024drivevlm, wang2024omnidrive}, and virtual reality systems~\cite{yang2024holodeck, chandrasegaran2024hourvideo}, current models still suffer from significant performance gaps. 
Enhancing the spatial reasoning capabilities of VLMs represents a critical frontier for real-world artificial intelligence applications\cite{zheng2025multimodal}.

% Despite these advancements, current models continue to exhibit suboptimal performance in spatial reasoning~\cite{liu2023visual, tong2024cambrian, yang2025thinking}. 
% The acquisition of this capability is essential for facilitating critical downstream applications, such as robot navigation~\cite{liu2023aerialvln, yokoyama2024vlfm}, robot manipulation~\cite{driess2022learning, huang2024rekep}, autonomous driving~\cite{tian2024drivevlm, wang2024omnidrive} and virtual reality systems~\cite{yang2024holodeck}. 
% The prevailing limitations of existing VLMs in spatial reasoning severely restrict the broader application of these models in such complex scenarios.

% SFT
To enhance the spatial reasoning capabilities of VLMs, early data-driven approaches based on Supervised Fine-Tuning (SFT) typically rely on post-training on massive synthetic datasets with auxiliary spatial representations~\cite{chen2024spatialvlm, cheng2024spatialrgpt, cai2025spatialbot, ma2025spatialllm}. However, these works are highly data-intensive, requiring substantial effort for data curation. Moreover, training via SFT tends to induce strict rote memorization, which ultimately restricts the models' generalization capabilities\cite{chu2025sft}.

% RL
To develop robust and generalizable spatial reasoning capabilities, recent advances in Reinforcement Learning with Verifiable Rewards (RLVR) are introduced to the spatial reasoning domain \cite{ma2025spatialreasoner, li2025spatialladder, ouyang2025spacer, wu2025spatial, zhan2025actial}. These works leverage Chain-of-Thought (CoT) prompting alongside verifiable outcome rewards to improve reasoning performance. 
However, relying exclusively on sparse outcome rewards hinders fine-grained advantage estimation, leading to inaccurate credit assignment for intermediate reasoning steps.
Concurrently, another line of research proposes structured CoT templates, aiming to improve perception in human-like cognitive process \cite{ma2026thinking} and enhance the interpretability of the reasoning trajectories \cite{batra2025spatialthinker, huang20253d}.
Nevertheless, these studies critically overlook depth-related information within their templates, fundamentally restricting the comprehension of 3D spaces. 
Furthermore, they also suffer from poor credit assignment, failing to reward specific reasoning behaviors in the structured information.

To address these critical gaps, we propose a structured reasoning CoT framework that explicitly models 3D spatial perception, ensuring robust environmental understanding. As illustrated in \cref{fig:teaser} (a), during inference, the model first extracts 3D information from the given image, followed by logical deduction grounded in the perceived spatial observations. 
This reasoning approach offers three primary advantages: 
First, it significantly improves the interpretability of the reasoning process, facilitating precise information extraction. 
Second, explicit modeling of spatial attributes ensures that the acquired reasoning capabilities are fundamentally grounded in physical reality.
More importantly, by partitioning the CoT into independently extractable perception and analysis modules, it unlocks the potential for fine-grained, targeted verification and accurate credit assignment for each reasoning step during RL training. 

\begin{figure*}[tb]
  \centering
  \includegraphics[width=\textwidth]{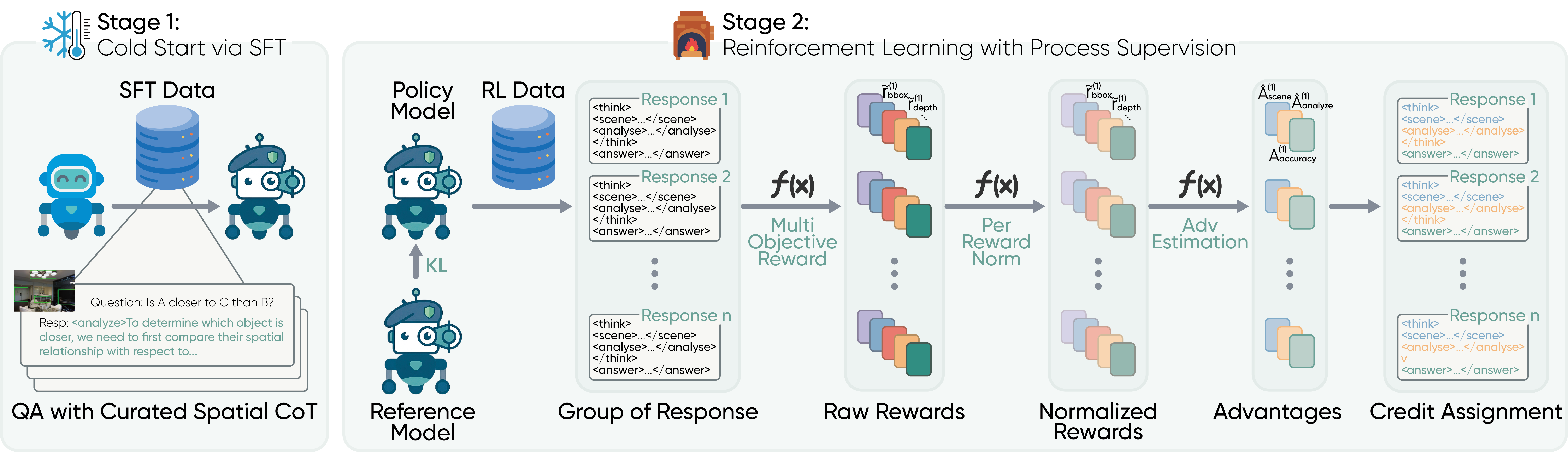}
  \caption{
  \textbf{Method overview of SCOUT.} Stage 1 initializes the training through a SFT cold-start using our curated CoT data. Stage 2 employs reinforcement learning with multi-objective rewards and fine-grained advantage estimation to effectively assign credit across different tokens and update the policy model.
  }
  \label{fig:method}
\end{figure*}

Building on the structured CoT, we introduce a novel reinforcement learning algorithm that integrates multi-objective process rewards coupled with an advanced advantage estimation method. 
Specifically, we design dedicated reward functions for object's bounding box(bbox) and depth perception, providing precise supervisory signals to optimize the model's 3D spatial understanding capabilities. 
Furthermore, we propose a reasoning consistency reward that evaluates the alignment between the reasoning trajectory and the final answer, effectively guiding the model to generate coherent deductions.
Crucially, the proposed advantage estimation algorithm integrates these process rewards to facilitate fine-grained credit assignment across different segments of the CoT trace. 
By explicitly attributing the contribution of each reasoning module to the final outcome, our RL algorithm fundamentally fosters advanced spatial reasoning in VLMs.

To support this framework, we develop a scalable data synthesis pipeline and introduce SCOUT-24k, a comprehensive structured CoT dataset. This dataset encompasses diverse spatial reasoning tasks, spanning the entire spectrum from fundamental relationship comprehension and relative distance prediction to complex perspective transformations.

Leveraging the proposed methodology and the SCOUT-24k dataset, we present SCOUT-3B and SCOUT-7B, initialized from Qwen2.5-VL-3B and Qwen2.5-VL-7B\cite{bai2025qwen2}, respectively. As illustrated in \cref{fig:teaser}(b), SCOUT-3B achieves substantial gains over the baseline models, yielding improvements of 16.85\% on general spatial benchmarks and 6.3\% on complex spatial reasoning tasks.
Consistent trends are observed upon scaling the model to 7B. Notably, SCOUT-7B further achieves an average accuracy improvement of 4.28\% and 0.87\% over GPT-4o\cite{hurst2024gpt} on two types of spatial reasoning benchmarks, respectively.
More importantly, despite being trained exclusively on single-image, SCOUT extends its spatial reasoning proficiency to multi-image and video domains. This is evidenced by performance gains of 2.46\% on ViewSpatial\cite{li2025viewspatial} dataset and 3.13\% on the multiple-choice section of VSI-Bench\cite{yang2025thinking}.
Collectively, these results establish SCOUT as a crucial step toward the next generation of spatially grounded VLMs.

Our main contributions can be summarized as follows:
\begin{enumerate}
    \item We propose a structured reasoning CoT framework that explicitly models the perception of 3D spatial environments, ensuring comprehensive spatial understanding and robust reasoning.
    % thereby significantly improving interpretability and ensuring that reasoning capabilities are fundamentally grounded in physical reality. 
    \item We introduce a novel RL algorithm integrating multi-objective process rewards with advanced advantage estimation, facilitating fine-grained credit assignment across different segments of the reasoning trajectory. 
    \item We develop a scalable data synthesis pipeline and construct the SCOUT-24k dataset, systematically spanning spatial reasoning tasks from fundamental relationship comprehension to complex perspective transformations.
    \item Extensive experiments demonstrate that SCOUT significantly outperforms baselines and GPT-4o in both general spatial benchmarks and complex spatial reasoning tasks, while exhibiting robust generalization to multi-image and video domain, validating the efficacy of our approach.
\end{enumerate}

\section{Method}
% In this section, we present the algorithm and training pipeline of SCOUT. We enhance spatial reasoning through two synergistic components: (1) a novel reinforcement learning framework featuring multi-objective process rewards and fine-grained advantage estimation to supervise the entire reasoning trajectory; and (2) a robust data synthesis pipeline to curate SCOUT-24k, a structured reasoning CoT dataset that explicitly incorporates depth-aware spatial information. To facilitate advanced spatial reasoning, our training method adopts a two-stage pipeline. The first stage serves as a supervised cold start utilizing our proposed SCOUT-24k dataset. This initial step aligns the model with the expected reasoning formats and structured outputs, thereby significantly accelerating the convergence of the subsequent RL phase. In the second stage, we deploy our core RL algorithm. As illustrated in Figure~\ref{fig:method}, this framework is driven by multi-objective process rewards and a fine-grained advantage estimation mechanism.

In this section, we present the algorithmic and data construction pipeline of SCOUT, as illustrated in \cref{fig:method} and \cref{fig:dataset-construction}. 
To equip VLMs with advanced spatial reasoning capabilities, our methodology integrates two essential components: a structured CoT template to facilitate robust 3D understanding and a novel RL algorithm that effectively incentivizes spatial reasoning within the structured CoT. 
For data curation, we create a scalable data synthesis pipeline, generating a comprehensive dataset of spatial reasoning tasks.
% Our training approach follows a dedicated designed two-stage pipeline. The first stage serves as a supervised cold start utilizing SCOUT-24k our curated structured CoT dataset that explicitly incorporates depth-aware spatial information. This initial step critically aligns the model with the expected reasoning formats and structured outputs, thereby significantly accelerating the convergence of the subsequent RL phase. In the second stage, we deploy our core RL algorithm. To accurately supervise the entire reasoning trajectory, this phase is uniquely driven by multi-objective process rewards and a fine-grained advantage estimation mechanism.

\subsection{Structured Reasoning Process}
We propose a novel structured CoT framework to enhance accurate physical perception of 3D environments in complex spatial reasoning tasks. The entire reasoning trajectory is encapsulated within a \texttt{<think>} block, which strictly guides the model through a progressive, modular cognitive pipeline before yielding the final answer. We initiates this reasoning process with the \texttt{<caption>} module, where the model generates a global semantic description of the visual input. Following this, the model transitions to the \texttt{<scene>} module, which quantitatively describe key objects with their bounding boxes and depths. Based on this fine-grained 3D spatial perception, the model then enters the \texttt{<analyze>} phase to perform explicit logical reasoning and deduction, seamlessly executing numerical comparisons and relative reasoning based on the extracted depth values. After completing comprehensive visual extraction and rigorous logical formulation, the model exits the \texttt{</think>} wrapper and outputs the final conclusion strictly within the \texttt{<answer>} tag. In summary, our proposed reasoning template adopts the following sequential format: \texttt{<think>} \texttt{<caption>}...\texttt{</caption>} \texttt{<scene>} ... \texttt{</scene>} \texttt{<analyze>} ... \texttt{</analyze>} \texttt{</think>} \texttt{<answer>} ... \texttt{</answer>}.

% As illustrated in Figure~\ref{fig:method}, we perform a brief SFT warm-up in the first stage of training to condition the model to generate the expected structured reasoning formats, thereby providing a valid and efficient starting point for the subsequent RL training.

As illustrated in \cref{fig:method}, we perform a preliminary SFT cold-start to guide the model for structured reasoning formats, establishing a stable foundation for effective policy optimization in the subsequent RL phase.

% significantly accelerating convergence in the subsequent RL phase.

\subsection{Multi-Objective Process Rewards}

% To facilitate advanced spatial reasoning, our training approach follows a two-stage procedure. The first stage serves as a supervised cold start using our proposed SCOUT-24k dataset. This step aligns the model with the expected reasoning format and structural outputs, significantly accelerating the convergence of the subsequent RL phase. The second stage deploys our core RL algorithm. As illustrated in Figure~\ref{fig:method}, this framework features multi-objective process rewards and a fine-grained advantage estimation mechanism. We start by introducing our reward design in detail.

To facilitate effective RL training, we design multiple process rewards for the verification of intermediate reasoning steps within our proposed structured CoT. Concretely, this involves 5 distinct rewards for a multi-objective process supervision of the reasoning process:

\subsubsection{Regularized Grounding Reward.}
To evaluate model's perception of environment, we first align the predicted objects $\mathcal{O}^{\text{pred}}$ with the ground-truth objects $\mathcal{O}^{\text{gt}}$ via the Hungarian algorithm\cite{kuhn1955hungarian}. The matching cost $\mathcal{C}_{i,j}$ between the $i$-th prediction and the $j$-th ground truth is:
\begin{equation}
  \mathcal{C}_{i,j} = \lambda_{\text{sem}}(1 - \text{sim}(l_i, l_j)) + \lambda_{\text{iou}}(1 - \text{EIoU}(b_i, b_j)) + \lambda_{\text{dep}}(1 - \delta(d_i, d_j)),
\end{equation}
where $l$, $b$, and $d$ denote the semantic label, bounding box, and depth, respectively. Here $\text{sim}(\cdot)$ is the semantic similarity, $\text{EIoU}(\cdot)$ is the Efficient IoU~\cite{zhang2022focal}, and $\delta(d_i, d_j) = \exp\left( -2 \frac{|d_i - d_j|}{d_j} \right)$ measures depth consistency. We prioritize geometric localization by setting the coefficients to $\lambda_{\text{sem}}=2.0$, $\lambda_{\text{iou}}=3.0$, and $\lambda_{\text{dep}}=0.5$. 

To mitigate reward hacking (e.g., bbox proliferation) seen in prior work~\cite{batra2025spatialthinker}, we introduce a cardinality penalty. The final regularized grounding reward is:
\begin{equation}
  r_{\text{grounding}} = \underbrace{\frac{1}{|\mathcal{M}|} \sum_{(i,j) \in \mathcal{M}} \text{EIoU}(b_i, b_j)}_{\text{Quality Term}} - \underbrace{\eta \cdot \max(0, |\mathcal{O}^{\text{pred}}| - |\mathcal{O}^{\text{gt}}|)}_{\text{Over-generation Penalty}},
\end{equation}
where $\mathcal{M}$ is the set of matched pairs and the penalty coefficient $\eta$ is set to $0.2$.

\subsubsection{Depth Reward.} 
Based on the matched pairs $\mathcal{M}$ obtained above, we impose a continuous Depth reward to penalize deviations in the $z$-axis: 
\begin{equation}
  r_{\text{depth}} = \frac{1}{|\mathcal{M}|} \sum_{(i,j) \in \mathcal{M}} \delta(d_i, d_j).
\end{equation}
This reward incentivizes the model's spatial understanding to 3D environment.

\subsubsection{Reasoning Consistency Reward.}
To prevent models from arriving at correct answers through flawed or disconnected reasoning, we employ a blind verification mechanism. We feed only the textual question and the generated reasoning chain into base model, omitting all visual inputs. The reward $r_{\text{consistency}}$ is 1 if the verifier derives the ground truth solely from this textual input, and 0 otherwise. This ensures the reasoning path logically entails the final prediction.

\subsubsection{Accuracy Reward.} 
This is the standard outcome-based reward. Given the multiple-choice format, we define a binary reward $r_{\text{accuracy}} \in \{0, 1\}$, where $r_{\text{accuracy}} = 1$ if the prediction matches the ground-truth answer, and $0$ otherwise.

\subsubsection{Format Reward.} 
Finally, to encourage format following, we assign a binary reward $r_{\text{format}} = 1$ if the model correctly encloses its reasoning process, spatial perception, and final output within the \texttt{<think>}, \texttt{<scene>}, and \texttt{<answer>} tags, respectively; otherwise, $r_{\text{format}} = 0$.

\subsection{Advantage Estimation for Fine-grained Credit Assignment}

Standard Group Relative Policy Optimization (GRPO)\cite{shao2024deepseekmath} assigns a single reward and advantage to the entire response sequence, suffering from coarse credit assignment across different reasoning steps. To address this, we propose a fine-grained Advantage Estimation mechanism based on our multi-objective rewards. Leveraging the explicit tags of our structured CoT, we assign distinct credit to three functional segments:  Perception (from \texttt{<think>} to \texttt{</scene>}), Analysis (from \texttt{<analyze>} to \texttt{</think>}), and Answer (from \texttt{<answer>} to \texttt{</answer>}.)

\subsubsection{Advantage Aggregation and Mixing.}

For a group of generated outputs, we first apply standard z-score normalization to each raw reward $r_k$ to yield $\tilde{r}_k$, mitigating scale discrepancies between heterogeneous rewards:
\begin{equation}
    \tilde{r}_k^{(i)} = \frac{r_k^{(i)} - \mu_k}{\sigma_k},
\end{equation}
where $r_k^{(i)}$ represents the $k$-th reward term for the $i$-th sample, and $\mu_k$ and $\sigma_k$ denote the group mean and standard deviation, respectively. To align with the distinct functions of each CoT stage, we aggregate these normalized rewards into stage-specific base advantages:
\begin{equation}
    \begin{aligned}
        \mathcal{A}_{\text{scene}} &= \tilde{r}_{\text{grounding}} + \tilde{r}_{\text{depth}}, \\
        \mathcal{A}_{\text{analyze}} &= \tilde{r}_{\text{consistency}}, \\
        \mathcal{A}_{\text{outcome}} &= \tilde{r}_{\text{format}} + \tilde{r}_{\text{acc}}.
    \end{aligned}
\end{equation}

A strictly decoupled reward might incentivize models to optimize intermediate steps while ignoring the correctness of final answer. Therefore, we blend local process supervision with the global outcome advantage:
\begin{equation}
    \begin{aligned}
        \hat{A}_{\text{scene}} &= \alpha_1 \mathcal{A}_{\text{scene}} + (1-\alpha_1) \mathcal{A}_{\text{outcome}}, \\
        \hat{A}_{\text{analyze}} &= \alpha_2 \mathcal{A}_{\text{analyze}} + (1-\alpha_2) \mathcal{A}_{\text{outcome}}.
    \end{aligned}
\end{equation}
% TODO:润色
Here, parameters $\alpha_1, \alpha_2 \in [0, 1]$ control the trade-off between local process alignment and global task success. We set $\alpha_1=0.3$ and $\alpha_2=0.3$ to ensure intermediate verification is strongly anchored by the final prediction accuracy.

\subsubsection{Token-Level Credit Assignment and Optimization.}

Finally, we allocate these advantages to specific tokens based on their functional segments. Let $T_{\text{scene}}$ and $T_{\text{think}}$ denote the indices of the \texttt{</scene>} and \texttt{</think>} tags, respectively. The token-wise advantage $A_t$ is defined as:

\begin{equation}
    A_t = 
    \begin{cases} 
        \hat{A}_{\text{scene}} & \text{if } t \le T_{\text{scene}}, \\
        \hat{A}_{\text{analyze}} & \text{if } T_{\text{scene}} < t \le T_{\text{think}}, \\
        \mathcal{A}_{\text{outcome}} & \text{if } t > T_{\text{think}}.
    \end{cases}
\end{equation}
This strategy ensures that the policy gradients for perception tokens are primarily driven by visual grounding and depth perception accuracy, while  those for reasoning tokens are governed by logical consistency. Crucially, both remain strictly anchored to the final answer accuracy. Finally, we optimize the policy parameters $\theta$ using a clipped surrogate objective with KL regularization:
\begin{equation}
    \mathcal{J}(\theta) = \mathbb{E}_{t} \left[ \min \left( \rho_t(\theta) A_t, \text{clip}(\rho_t(\theta), 1 - \epsilon, 1 + \epsilon) A_t \right) \right] - \beta D_{\text{KL}}(\pi_\theta \| \pi_{\text{ref}}),
\end{equation}
where $\rho_t(\theta) = \frac{\pi_\theta(y_t | x, y_{<t})}{\pi_{\text{old}}(y_t | x, y_{<t})}$ denotes the probability ratio between the current policy and the old policy at time step $t$, and $\beta$ serves as an adaptive coefficient controlling the KL divergence to ensure that the policy does not deviate excessively from the reference model $\pi_{\text{ref}}$.

\begin{figure*}[tb]
  \centering
  \includegraphics[width=\textwidth]{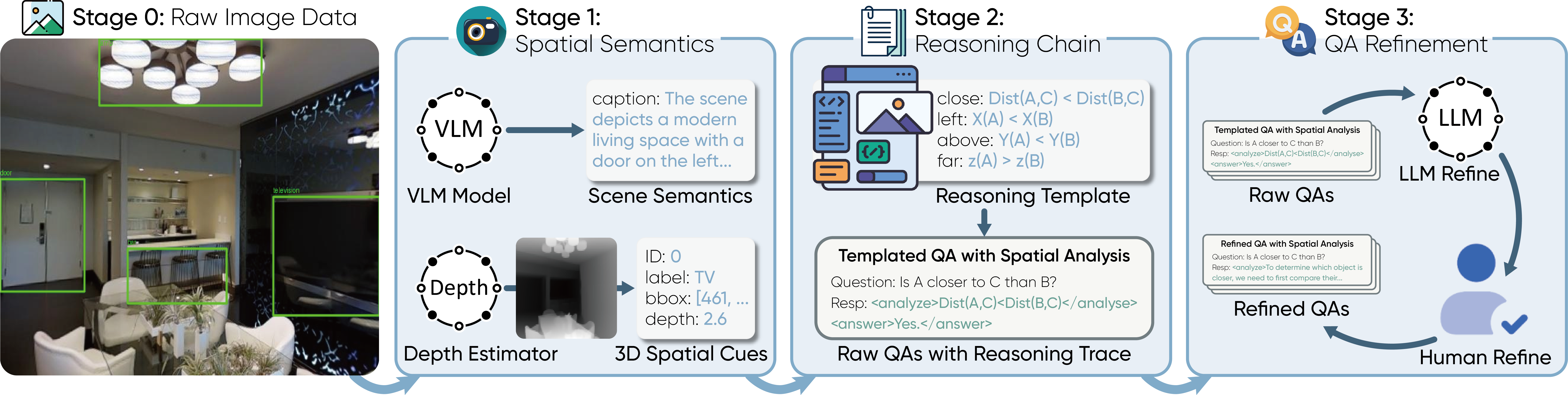}
  \caption{
  \textbf{Overview of SCOUT-24k dataset construction pipeline} from raw images to final spatial reasoning data generation with representative Relative Distance task.
  }
  \label{fig:dataset-construction}
\end{figure*}

\subsection{Data Construction}
To facilitate SFT and RL training of the proposed structured reasoning framework, high-quality training data for perception and reasoning is essential. We introduce SCOUT-24k, which contains four types of structured spatial reasoning data. \Cref{fig:dataset-construction} illustrates our construction pipeline.

\subsubsection{Task Design and Categories.} 
% The meticulously designed dataset encompasses four task categories. These include spatial relation understanding, relative distance prediction, and perspective transformation reasoning. The fourth category covers tasks distinct from the first three, such as counting, object size estimation, existence verification and others.
% dataset 包含4中spaital related QAs. (1) Spatial relation understanding: 这种QA要求模型回答多种物品之间的相对空间关系。(2) Relative distance prediction: 这类QA要求模型预测物品之间的相对远近。(3) Perspective transformation reasoning: 这类QA主要考察在观测视角发生变化时，模型是否能够推测场景物品相对空间关系的变化，(4) Object-centric spatial reasoning: 这类QA主要考察模型对于物品空间属性的推理，例如物品的尺寸和是否存在等。
The dataset comprises four categories of spatially related QAs:
(1) Spatial Relation Understanding, which requires the model to identify and reason about the relative spatial relationships among multiple objects;
(2) Relative Distance Prediction, where the model is asked to determine the relative proximity between objects;
(3) Perspective Transformation Reasoning, which evaluates whether the model can infer changes in relative spatial relationships when the viewpoint shifts; and
(4) Object-Centric Spatial Reasoning, which assesses the model’s ability to reason about object-level spatial attributes, such as size and existence.

\subsubsection{Construction Pipeline.} 
An overview of our construction pipeline is illustrated in \cref{fig:dataset-construction}. Source images are collected from EmbSpatial~\cite{du2024embspatial} and STVQA~\cite{batra2025spatialthinker} datasets, both of which provide detailed bounding box annotations and object labels.

In the first stage, we extract spatial and semantic information from each image. We leverage Qwen-VL-Max together with a state-of-the-art monocular depth estimator, Depth-Anything-3~\cite{depthanything3}, to generate scene captions and estimate the center coordinates $(u, v, z)$ for each object. The depth is sampled at the center of the corresponding bounding box.
This process yields a JSON-formatted structured scene representation, which serves as the perception ground truth in the subsequent chain-of-thought reasoning.

In the second stage, we construct the textual reasoning process. Using the derived 3D spatial information, we synthesize reasoning steps and corresponding answers for relative distance prediction and perspective transformation reasoning tasks, based on templates adapted from CV-Bench~\cite{tong2024cambrian} and Spatial-SSRL~\cite{liu2025spatial}.

Finally, we refine the generated reasoning processes. We prompt Qwen-VL-Max~\cite{bai2025qwen2} with the scene caption and synthesized reasoning steps, and retain only those samples whose predicted answers match the ground-truth labels. Subsequently, expert human annotators further review and refine the selected samples to ensure quality and consistency.

Notably, data from STVQA and EmbSpatial already contain answer annotations. For these samples, we execute only the first stage of the pipeline and reserve them for reinforcement learning. Additional details regarding dataset construction are provided in the supplementary material.

\section{EXPERIMENTS}

\begin{table}[t]
\centering
\caption{\textbf{Performance comparison on general spatial benchmarks.} We include SCOUT in the comparison groups. The best results in each comparison group are highlighted in \textbf{bold}, and the second-best results are \underline{underlined}.}
\label{tab:in_domain_spatial_performance}

% 2. 稍微缩减列间距

% 3. 压缩行高，让表格在垂直方向更紧凑
% \renewcommand{\arraystretch}{0.95}

% 将表格宽度匹配栏宽
\resizebox{\linewidth}{!}{%
\begin{tabular}{l|c|ccc|c|c} % 
\toprule
% --- 表头 ---
\multirow{2}{*}{\textbf{Model}} & \multirow{2}{*}{\textbf{EmbSpatial}} & \multicolumn{3}{c|}{\textbf{CV-Bench}} & \textbf{BLINK} & \multirow{2}{*}{\textbf{Overall}} \\ 
 &  & \textbf{2D} & \textbf{3D} & \textbf{Avg} & \textbf{(Depth)} & \\ 
\midrule

% --- Group 1: Proprietary Models ---
\rowcolor{rowgray} 
\multicolumn{7}{c}{\textit{Proprietary Models}} \\ % >>> 6 改为 7 <<<
GPT-4o & 70.27 & 75.44 & 83.08 & 79.26 & 76.61 & 75.38 \\
\midrule

% --- Group 2: Open source and specialized spatial models ---
\rowcolor{rowgray} 
\multicolumn{7}{c}{\textit{Open Source and Specialized Spatial Models}} \\ % >>> 6 改为 7 <<<
Intern-VL3.5-4B & 70.96 & 70.30 & 79.75 & 75.03 & 62.90 & 69.63 \\
Intern-VL3.5-8B & 74.86 & 77.05 & 82.91 & 79.98 & 63.70 & 72.85 \\
SpaceLLaVA & 47.85 & 59.89 & 62.83 & 61.36 & 57.25 & 55.49 \\
SpatialBot & 51.67 & 66.17 & 67.16 & 66.67 & 59.67 & 59.34 \\
\midrule

% --- Group 3: Qwen2.5-VL-3B Based ---
\rowcolor{rowgray} 
\multicolumn{7}{c}{\textit{Spatial Models Based on Qwen2.5-VL-3B}} \\ % >>> 6 改为 7 <<<
Qwen2.5-VL-3B & 59.86 & 66.55 & 63.50 & 65.03 & 57.25 & 60.71 \\
SpaceOm-3B & 61.07 & 71.27 & 71.50 & 71.39 & 57.25 & 63.24 \\
SpatialLadder-3B & 58.10 & 68.91 & 70.41 & 69.66 & \underline{66.12} & 64.63 \\
SpatialThinker-3B & \underline{61.89} & \underline{73.01} & \underline{76.16} & \underline{74.59} & 62.90 & \underline{66.46} \\
% >>> 高亮 SCOUT-3B 行 <<<
\rowcolor{rowgreen}
\textbf{SCOUT-3B (Ours)} & \textbf{76.95} & \textbf{75.73} & \textbf{80.91} & \textbf{78.32} & \textbf{77.42} & \textbf{77.56} \\ 
\midrule

% --- Group 4: Qwen2.5-VL-7B Based ---
\rowcolor{rowgray}
\multicolumn{7}{c}{\textit{Spatial Models Based on Qwen2.5-VL-7B}} \\ % >>> 6 改为 7 <<<
Qwen2.5-VL-7B & 59.91 & 74.89 & 79.41 & 77.15 & 73.38 & 70.15 \\
SpatialReasoner-7B & 47.30 & 47.35 & 76.58 & 61.97 & \underline{77.42} & 62.23 \\
SpatialThinker-7B & \underline{63.35} & \underline{75.59} & \underline{81.83} & \underline{78.71} & 72.58 & \underline{71.55} \\
% >>> 高亮 SCOUT-7B 行 <<<
\rowcolor{rowgreen}
\textbf{SCOUT-7B (Ours)} & \textbf{78.76} & \textbf{78.85} & \textbf{83.50} & \textbf{81.18} & \textbf{79.03} & \textbf{79.66} \\ 
\bottomrule
\end{tabular}
}
\end{table}

\subsection{Experimental Settings}

\subsubsection{Implementation Details.}
We train our SCOUT models based on the Qwen2.5-VL series models~\cite{bai2025qwen2} (3B and 7B). Our training pipeline consists of two sequential stages:
(1) \textbf{SFT Cold-Start:} We apply Low-Rank Adaptation (LoRA) \cite{hu2022lora} to all linear modules with a rank of $r=8$. The model is fine-tuned for one epoch on the SFT subset of SCOUT-24k using a learning rate of $1 \times 10^{-4}$.
(2) \textbf{RL Training:} We train the model using our proposed RL algorithm for 200 steps, with a global batch size of 128 and a learning rate of $1 \times 10^{-6}$. To stabilize policy updates, we enforce KL divergence penalty with a coefficient of $\beta=0.01$. During exploration, we sample $N=8$ trajectories per prompt at a temperature of $T=1.0$. Checkpoints are saved every 50 steps, and we report the performance of the best-performing checkpoint. Complete implementation details can be found in the supplementary material.

\subsubsection{Evaluation Setup.}
We compare SCOUT against a comprehensive suite of baselines, categorized into four distinct groups:
(i) \textbf{Proprietary Models}, such as GPT-4o~\cite{hurst2024gpt};
(ii) \textbf{Open-Source Models}, such as the Intern-VL3.5 series~\cite{wang2025internvl3_5};
(iii) \textbf{Spatial Specialist Models}, including SpatialBot~\cite{cai2025spatialbot} and SpaceLLaVA~\cite{remyxai2024SpaceLLaVA} (a public reimplementation of SpatialVLM~\cite{chen2024spatialvlm}); and
(iv) \textbf{Spatial Models Based on the Qwen2.5-VL Series}~\cite{bai2025qwen2}, encompassing variants built upon both the 3B and 7B architectures (e.g., SpaceOm~\cite{remyxai2025spaceom}, SpatialLadder~\cite{li2025spatialladder}, SpatialReasoner~\cite{ma2025spatialreasoner}, and SpatialThinker~\cite{batra2025spatialthinker}).

% LLaVA-OV-1.5 series~\cite{LLaVA-OneVision-1.5}, SpaceThinker~\cite{remyxai2025spacethinker}, SpaceQwen~\cite{remyxai2025spaceom}

Evaluations are conducted on a comprehensive suite of six single-image benchmarks, ranging from general spatial understanding to complex spatial reasoning. Specifically, we utilize EmbSpatial~\cite{du2024embspatial}, CV-Bench~\cite{tong2024cambrian}, and BLINK~\cite{fu2024blink} for general spatial evaluation while employing RoboSpatial~\cite{song2025robospatial}, SpatialBench~\cite{cai2025spatialbot}, and 3DSRBench~\cite{ma20253dsrbench} for complex spatial reasoning tasks. These benchmarks collectively assess fine-grained spatial reasoning capabilities, including depth estimation, relative distance reasoning, and perspective-dependent understanding. Furthermore, to verify our model's generalization to multi-image and video scenarios, we report the performance achieved on ViewSpatial~\cite{li2025viewspatial} and VSI-Bench~\cite{yang2025thinking}.

We adopt a strict zero-shot evaluation protocol to assess the models' intrinsic capabilities. To ensure reproducibility, inference is conducted using greedy decoding (temperature $T=0.0$). For a fair comparison, we apply model-specific CoT instructions for reasoning-oriented models where applicable. Accuracy serves as the primary evaluation metric across all tasks. Our evaluation pipeline extends the codebase of SpatialThinker~\cite{batra2025spatialthinker}, adapting it to support diverse input formats. Further evaluation details are available in the supplementary material.

\begin{table}[t]
\caption{\textbf{Performance comparison on complex spatial reasoning benchmarks.} SCOUT are included for comparison. The best results in each group are highlighted in \textbf{bold}, and the second-best results are \underline{underlined}.}
\label{tab:complex_spatial_performance}
\centering

\resizebox{\linewidth}{!}{%
\begin{tabular}{l|c|c|c|c} % >>> 增加了1个c <<<
\toprule
% --- 表头 ---
\textbf{Model} & \textbf{RoboSpatial} & \textbf{SpatialBench} & \textbf{3DSRBench} & \textbf{Overall} \\ 
\midrule

% --- Group 1: Proprietary Models ---
\rowcolor{rowgray} 
\multicolumn{5}{c}{\textit{Proprietary Models}} \\ % >>> 4改为5 <<<
GPT-4o &  70.17 & 67.81 & 44.78 & 60.92 \\
\midrule

% --- Group 2: Open source and specialized spatial models ---
\rowcolor{rowgray}
\multicolumn{5}{c}{\textit{Open source and specialized spatial models}} \\ % >>> 4改为5 <<<
Intern-VL3.5-4B & 63.59 & 57.47 & 42.10 & 54.39 \\
Intern-VL3.5-8B & 74.12 & 64.94 & 42.06 & 60.37 \\
SpaceLLaVA & 47.36 & 36.78 & 42.02 & 42.05 \\
SpatialBot & 54.38 & 54.02 & 45.83 & 51.41 \\
\midrule

% --- Group 3: Qwen2.5-VL-3B Based ---
\rowcolor{rowgray}
\multicolumn{5}{c}{\textit{Spatial Models based on Qwen2.5-VL-3B}} \\ % >>> 4改为5 <<<
Qwen2.5-VL-3B & 59.64 & 55.74 & 40.65 & 52.01 \\
SpaceOm-3B & 69.29 & 57.47 & 41.07 & 55.94 \\
SpatialLadder-3B & 70.61 & 55.74 & 42.44 & 56.26 \\
SpatialThinker-3B & \underline{71.05} & \textbf{59.77} & \underline{43.31} & \underline{58.04} \\
% >>> 高亮 SCOUT-3B 行 <<<
\rowcolor{rowgreen} 
\textbf{SCOUT-3B (Ours)} & \textbf{72.81} & \underline{58.04} & \textbf{44.08} & \textbf{58.31} \\ 
\midrule

% --- Group 4: Qwen2.5-VL-7B Based ---
\rowcolor{rowgray} 
\multicolumn{5}{c}{\textit{Spatial Models based on Qwen2.5-VL-7B}} \\ % >>> 4改为5 <<<
Qwen2.5-VL-7B & 64.91 & 62.64 & 46.13 & 57.89 \\
SpatialReasoner-7B & 61.84 & 50.00 & \textbf{53.79} & 55.21 \\
SpatialThinker-7B & \textbf{74.12} & \underline{63.79} & 46.59 & \underline{61.50} \\
% >>> 高亮 SCOUT-7B 行 <<<
\rowcolor{rowgreen} 
\textbf{SCOUT-7B (Ours)} & \underline{71.05} & \textbf{66.09} & \underline{48.23} & \textbf{61.79} \\ 

\bottomrule
\end{tabular}
}
\end{table}

\subsection{Main Results}

% \Cref{tab:in_domain_spatial_performance} comprehensively summarizes the evaluation results on the general spatial benchmarks. SCOUT achieves new state-of-the-art performance at both the 3B and 7B scales, outperforming all baselines in their respective categories by significant margins of 16.85\% and 9.51\%. The performance gains are particularly pronounced on the EmbSpatial benchmark, where SCOUT-3B (76.95\%) and SCOUT-7B (78.76\%) surpass their strongest same-scale competitors by massive margins of over +15\%. Most remarkably, SCOUT-7B strictly outperforms the proprietary GPT-4o across all individual benchmarks. Impressively, even our smaller SCOUT-3B variant achieves a higher overall score (77.56) than GPT-4o (75.38), decisively demonstrating the effectiveness of our proposed method.

\Cref{tab:in_domain_spatial_performance} comprehensively summarizes the evaluation results on the general spatial benchmarks. SCOUT achieves new state-of-the-art performance at both the 3B and 7B scales, outperforming all baselines in their respective categories by significant margins of 16.85\% and 9.51\%. Most remarkably, SCOUT-7B strictly outperforms the proprietary GPT-4o across all individual benchmarks. Impressively, even our smaller SCOUT-3B variant achieves a higher overall score (77.56) than GPT-4o (75.38), decisively demonstrating the effectiveness of our proposed method.

\Cref{tab:complex_spatial_performance} further highlights SCOUT's capabilities in complex spatial reasoning. At the 3B scale, SCOUT achieves 72.81\% accuracy on RoboSpatial (+13.17\% over Qwen2.5-VL-3B) and secures the highest overall score (58.31) in its group. This advantage amplifies at the 7B scale. Remarkably, SCOUT-7B achieves the highest overall score (61.79) in the entire evaluation, surpassing even the proprietary GPT-4o (60.92). These results confirm SCOUT's profound capability to tackle highly complex spatial reasoning tasks rather than merely memorizing simple spatial patterns.

\begin{table}[t]
\caption{\textbf{Performance comparison on ViewSpatial (Multi-Image) and VSI-Bench (Video).} We compare the baseline Qwen2.5-VL with our SCOUT across different scales. NQ and MCQ for VSI-Bench denotes numerical question and multiple-choice question, respectively.}
\label{tab:video_mv_performance}
\centering

% 修正为 4 列: lccc
\setlength{\tabcolsep}{15pt}
\begin{tabular}{lccc}
\toprule
\multirow{2}{*}{\textbf{Model}} & \textbf{ViewSpatial} & \multicolumn{2}{c}{\textbf{VSI-Bench}} \\ 
\cmidrule(lr){2-2} \cmidrule(lr){3-4}
 & \textbf{Overall} & \textbf{NQ} & \textbf{MCQ} \\ 
\midrule

% --- Group 1: 3B Models ---
Qwen2.5-VL-3B & 36.20 & 25.03 & 31.65 \\
SCOUT-3B & 38.91 & 19.26 & 32.97 \\
\midrule
% --- Group 2: 7B Models ---
Qwen2.5-VL-7B & 38.03 & 40.52 & 34.94 \\
SCOUT-7B & 40.49 & 29.35 & 38.07 \\

\bottomrule
\end{tabular}
\end{table}

\subsection{Out-of-domain Evaluation.}

\Cref{tab:video_mv_performance} validates the out-of-domain generalization on spatial reasoning tasks in multi-image and video scenarios. Despite being trained exclusively on single image, SCOUT transfers robustly to these domains. On ViewSpatial benchmark, SCOUT achieves overall accuracies of 38.91\% (3B) and 40.49\% (7B), successfully surpassing their respective baselines. Similar robust transferability is observed in the multiple-choice questions of VSI-Bench across both scales. 
Notably, we observe a performance drop in the numerical questions of VSI-Bench. This indicates a significant gap between the single-image and video domain in terms of  temporal object tracking, requiring a more specialized set of spatial reasoning capabilities to perform absolute numerical estimation across dynamic scenes.

Nevertheless, the consistent gains across other metrics strongly demonstrate that the proposed training approach fundamentally enhances intrinsic structural spatial understanding, enabling generalization to diverse visual formats.
% While we observe a performance decline in the Numerical Questions, this is an expected outcome: our current training corpus primarily focuses on relative spatial relationships rather than absolute measurement tasks that require direct numerical responses. 

\begin{table}[t]
\caption{\textbf{Ablation study on reward configurations.} We compare the full SCOUT-3B model against several baseline variants: Supervised Fine-Tuning (SFT), GRPO with vanilla Chain-of-Thought (GRPO + vanila CoT), GRPO without purposed multi-objective process rewards (w/o Process.), GRPO without fine-grained credit assignment (w/o Credit.), and our method omitting either the perception advantage ($\alpha_1=0$) or the reasoning advantage ($\alpha_2=0$). Across six spatial benchmarks, the best results are \textbf{bolded} and second-best are \underline{underlined}.}
\label{tab:ablation_study}
\centering

\resizebox{\textwidth}{!}{

\begin{tabular}{l|cccccc|c}
\toprule
\textbf{Method} & \textbf{EmbSpatial} & \textbf{CV-Bench} & \textbf{BLINK(Depth)} & \textbf{RoboSpatial} & \textbf{SpatialBench} & \textbf{3DSRBench} & \textbf{Avg.} \\ 
\midrule

% Row 1: SFT
SFT & 60.52 & 69.66 & 58.06 & 62.71 & 45.40 & 34.74 & 55.18 \\

% Row 2: GRPO + vanila CoT
GRPO + vanila CoT & \underline{76.07} & 75.63 & 65.32 & 69.73 & 54.02 & 42.63 & 63.90 \\

% Row 3: GRPO w/o Process.
GRPO w/o Process. & 74.01 & 76.02 & 70.96  & 68.85 & \textbf{58.62} & 42.44 & 65.15 \\

% Row 4: GRPO w/o Credit.
GRPO w/o Credit. & 75.00 & 75.90 & \underline{72.58} & \textbf{72.81} & 52.87 & 42.29 & 65.24 \\

% Row 5: GDPO alpha1 = 0
$\alpha_1 = 0$ & 74.25 & 77.55 & 71.77 & \underline{71.05} & 55.17 & \textbf{44.57} & \underline{65.73} \\

% Row 6: GDPO alpha2 = 0
$\alpha_2 = 0$ & 75.76 & \underline{78.11} & 71.77 & 67.54 & 55.74 & 42.25 & 65.20 \\

% Row 7: SCOUT (Ours)
SCOUT (Ours) & \textbf{76.95} & \textbf{78.32} & \textbf{77.42} & \textbf{72.81} & \underline{58.04} & \underline{44.08} & \textbf{67.94} \\

\bottomrule
\end{tabular}
}
\end{table}

\begin{figure*}[tb]
  \centering
  \includegraphics[width=\textwidth]{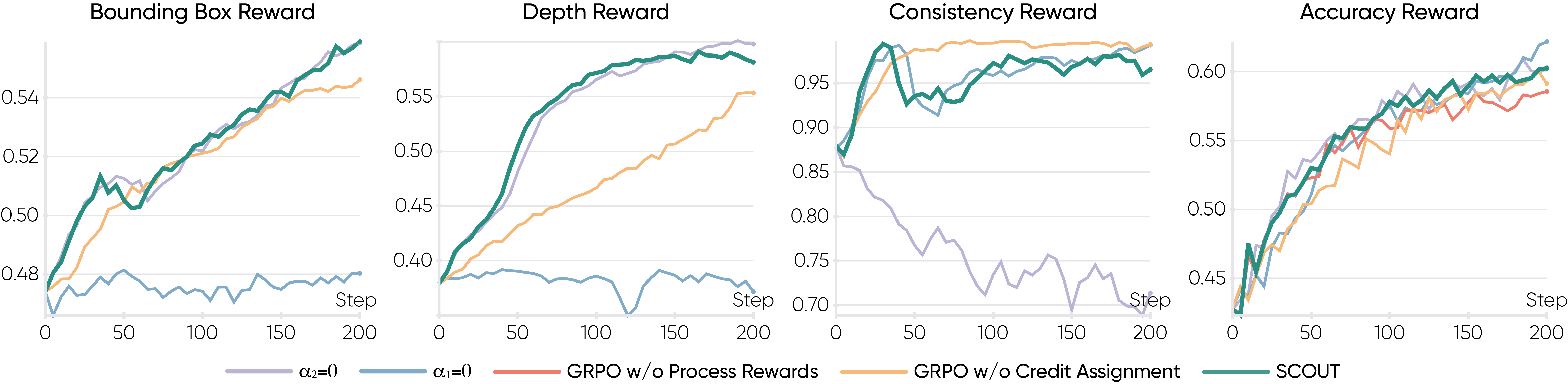}
  \caption{
    \textbf{Training dynamics under different reward configurations.} We track four specific reward components across training steps. Notably, ablating the perception advantage ($\alpha_1=0$) leads to a complete failure in optimizing Grounding and Depth reward. Conversely, excluding the reasoning advantage ($\alpha_2=0$) causes a severe collapse in the Consistency reward. SCOUT-3B effectively balances all components, achieving superior convergence and stability.
  }
  \label{fig:training curve}

\end{figure*}

\begin{table}[t]
\caption{\textbf{Sensitivity analysis of the factor $\alpha$.} We evaluate the performance of our model under different configurations of $\alpha_1 = \alpha_2 = \alpha \in \{0.3, 0.5, 0.7\}$. Here, $\alpha = 0.3$ corresponds to our default SCOUT-3B model. Across six spatial benchmarks, the best results are \textbf{bolded} and second-best are \underline{underlined}.}
\label{tab:sensitivity_alpha}
\centering

\resizebox{\textwidth}{!}{

\begin{tabular}{l|cccccc|c}
\toprule
\textbf{Method} & \textbf{EmbSpatial} & \textbf{CV-Bench} & \textbf{BLINK(Depth)} & \textbf{RoboSpatial} & \textbf{SpatialBench} & \textbf{3DSRBench} & \textbf{Avg.} \\ 
\midrule

% Row 1: alpha = 0.7
$\alpha = 0.7$ & 75.30 & \underline{77.86} & 75.00 & 70.61 & 55.74 & 42.59 & 66.18 \\

% Row 2: alpha = 0.5
$\alpha = 0.5$ & \underline{75.90} & 77.79 & \underline{75.80} & \underline{71.92} & \underline{56.32} & \underline{43.01} & \underline{66.79} \\

% Row 3: alpha = 0.3 (Ours)
$\alpha = 0.3$ (Ours) & \textbf{76.95} & \textbf{78.32} & \textbf{77.42} & \textbf{72.81} & \textbf{58.04} & \textbf{44.08} & \textbf{67.94} \\

\bottomrule
\end{tabular}
}
\end{table}

\subsection{Ablation Study}
We conduct ablation study at the 3B scale. \Cref{tab:ablation_study} and \cref{fig:training curve} validate the necessity of each reward component within our purposed approach. Overall, the full method achieves the highest average accuracy of 67.94\%, significantly outperforming standard GRPO baselines trained without process rewards (65.15\%) or without fine-grained credit assignment (65.24\%). This substantial gap indicates that standard policy optimization struggles to balance perception and reasoning without our proposed advantage estimating algorithm.

The importance of our fine-grained advantage formulation is clearly demonstrated by isolating specific components. As shown in \cref{fig:training curve}, when the perception advantage is excluded during optimization ($\alpha_1 = 0$), the grounding and depth reward fail to optimize entirely. This causes the overall average to drop to 65.73\%, severely impacting fundamental perception metrics like BLINK(Depth).

Conversely, excluding the reasoning advantage ($\alpha_2 = 0$) leads to a catastrophic collapse in the consistency reward during training. This removal causes severe overall performance degradation (dropping to 65.20\%), particularly impairing complex spatial tasks that rely on strict logical progression. For instance, accuracy on the RoboSpatial benchmark falls drastically from 72.81\% to 67.54\%.

We further conduct a sensitivity analysis for the factors by setting $\alpha_1 = \alpha_2$. As shown in \cref{tab:sensitivity_alpha}, $\alpha = 0.3$ achieves the best overall performance, while the average accuracy steadily degrades as $\alpha$ increases. This trend suggests that while process rewards are crucial for guiding step-by-step reasoning, overly emphasizing them may weaken the anchoring effect of the final correctness reward.

% Although some ablated variants achieve peak performance on isolated metrics (e.g., the $\alpha_2 = 0$ setting yields 78.11\% on CV-Bench, and $\alpha_1=0$ yields 44.57\% on 3DSRBench), this indicates an over-optimization for specific skills at the cost of broad generalization. Collectively, these findings confirm that integrating both perception and reasoning through our comprehensive reward design is essential for robust spatial understanding across diverse domains.

\subsection{Case Study}
\begin{figure*}[tb]
  \centering
  \includegraphics[width=\textwidth]{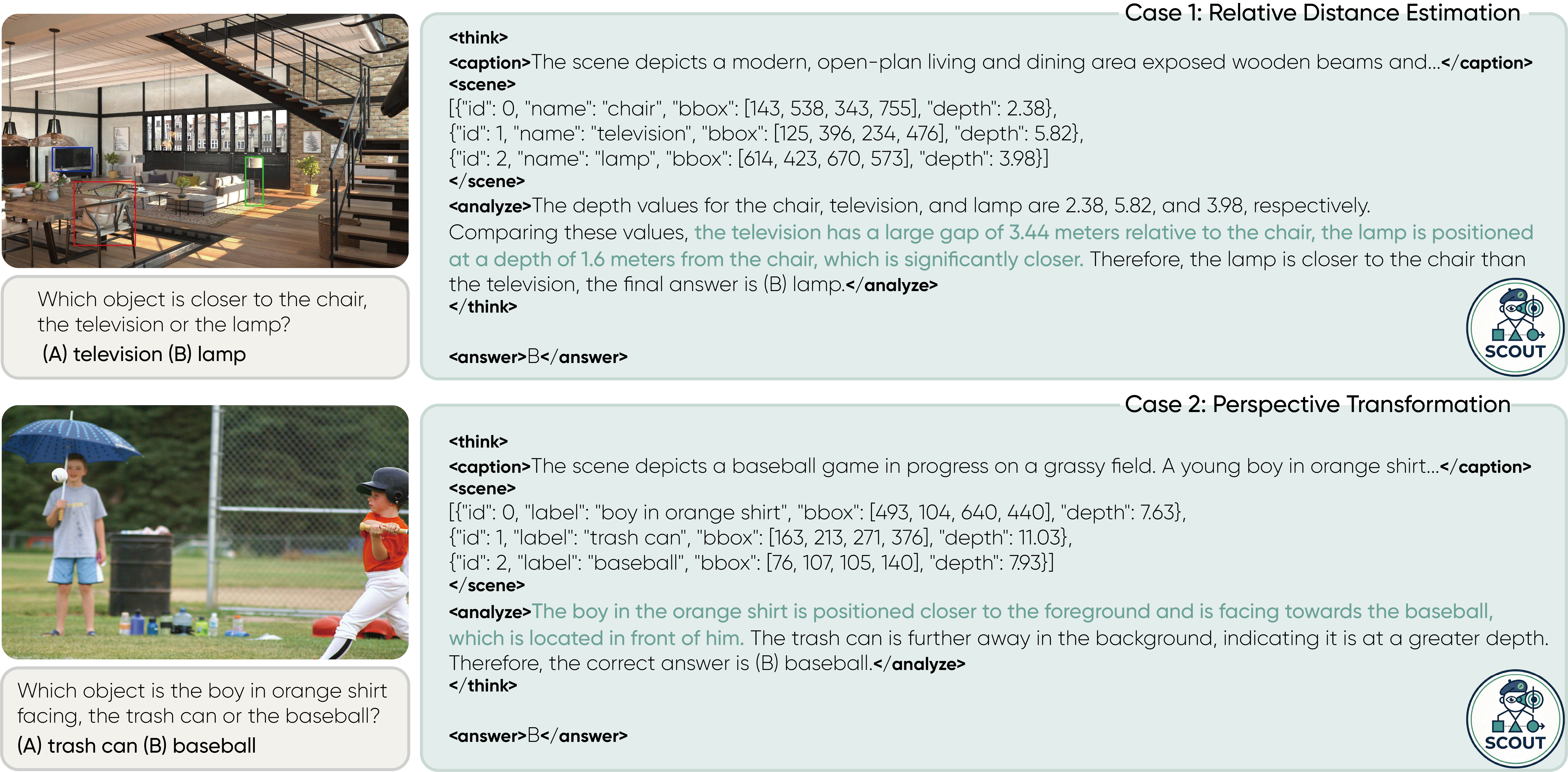}
  \caption{
  \textbf{Qualitative examples of the reasoning process in SCOUT.} SCOUT decomposes complex spatial reasoning into three steps: scene comprehension, fine-grained physical perception, and logical analysis.
  }
  \label{fig:case study}

\end{figure*}

Qualitative analysis reveals that SCOUT exhibits advanced analytical capabilities by developing structured spatial reasoning trajectories. \Cref{fig:case study} demonstrates that precise perception with 2D bounding box and 3D depth acts as a reliable foundation for generating logical reasoning chains.

As shown in Case 1, when handling relative distance estimation tasks, SCOUT systematically compares exact extracted depth metrics to resolve spatial ambiguities, avoiding reliance on mere 2D visual cues. Meanwhile, in Case 2, by integrating the visual orientation of objects with their perceived 3D spatial layout, the model successfully infers relative positions from the viewpoint of the boy. Ultimately, SCOUT consistently delivers accurate conclusions accompanied by transparent reasoning trajectories, proving that explicit structured CoT with 3D spatial perception is crucial for supporting robust reasoning.

% TODO: 这两段逻辑好像有点奇怪
\section{Related Work}

\subsection{Spatial Reasoning via Supervised Fine-Tuning}
Despite the success of VLMs in general vision-language tasks~\cite{comanici2025gemini, bai2025qwen2, deitke2025molmo, liu2025nvila, liu2023visual}, robust spatial reasoning remains a critical bottleneck. Existing SFT methods address this either via data-intensive post-training on synthetic datasets~\cite{chen2024spatialvlm, cheng2024spatialrgpt, cai2025spatialbot} or by integrating specialized geometric encoders~\cite{daxberger2025mm, ma2025spatialllm, liu2025ssr, hong20233d, huang2023embodied}. However, these approaches incur high data/computational overhead or limit architectural versatility. Crucially, SFT-based models tend to memorize training templates rather than genuinely generalizing spatial principles.

\subsection{Spatial Reasoning via Verifiable Reinforcement Learning}
To improve generalization, recent works adapt Reinforcement Learning with Verifiable Rewards (RLVR) to multimodal domains~\cite{guo2025deepseek, meng2025mm, yang2025r1, huang2025vision, liu2025visual, yu2025perception, shen2025vlm}, including spatial reasoning tasks and structured CoT design~\cite{ma2025spatialreasoner, li2025spatialladder, ouyang2025spacer, wu2025spatial, ma2026thinking, batra2025spatialthinker, huang20253d}. Nonetheless, current approaches face two key limitations: (1) their structured CoTs overlook critical depth-related information for 3D perception, and (2) their reliance on sparse outcome rewards hinders accurate credit assignment for intermediate steps. To resolve these, we propose a depth-aware structured CoT to capture 3D cues, coupled with multi-objective process rewards and a fine-grained advantage estimation algorithm to ensure precise credit assignment.

\section{Conclusion}
In this work, we present SCOUT, a novel model designed to address the critical bottleneck of VLMs in 3D spatial reasoning capabilities. By coupling a depth-aware structured CoT with a tailored process-supervised reinforcement learning algorithm, SCOUT effectively resolves the persistent challenges of sparse rewards and inaccurate credit assignment. Additionally, we introduce SCOUT-24k, a comprehensive dataset spanning from basic spatial relations to complex perspective transformations. Extensive evaluations demonstrate that the SCOUT-7B model achieves state-of-the-art performance from general spatial benchmarks to complex spatial reasoning tasks, while exhibiting robust out-of-domain generalization across multi-image and video scenarios. This research establishes a new approach for cultivating spatial reasoning in VLMs, paving a highly promising pathway toward the next generation of spatially aware AI systems.

%%%%%%%%%%%%%%%%%%% 上面这些部分，在结论之前的部分必须在14页以内。 %%%%%%%%%%%%%%%%%%% 

% \section{CheckList}
% \begin{itemize}
% \item I have removed all \verb| \vspace| and \verb|\hspace|  commands from my paper.
% \item I have not used \verb|\cite| command in the abstract.
% \item I have entered a correct \verb|\titlerunning{}| command and selected a meaningful short name for the paper.
% \item I have used the same name spelling in all my papers accepted to ECCV and ECCV Workshops.
% \item I have added acknowledgments without a section number, e.g. using the \verb|\section*{}| command.
% \item Excluding references and acknowledgments, my paper is no longer than 14 pages.
% \item I have not decreased the font size of any part of the paper (except tables) to fit into 14 pages, I understand Springer editors will remove such commands.
% \end{itemize}

% \section*{Acknowledgements}
% Please insert your acknowledgments here.


% \include{appendix}

\bibliographystyle{splncs04}
\bibliography{main}

\begin{thebibliography}{10}
\providecommand{\url}[1]{\texttt{#1}}
\providecommand{\urlprefix}{URL }
\providecommand{\doi}[1]{https://doi.org/#1}

\bibitem{remyxai2024SpaceLLaVA}
AI, R., Mayorquin, S.: Spacellava models. \url{https://huggingface.co/remyxai/SpaceLLaVA} (2024), accessed: 2026-02-03

\bibitem{remyxai2025spaceom}
AI, R., Mayorquin, S.: Spaceom models. \url{https://huggingface.co/remyxai/SpaceOm} (2025), accessed: 2026-02-03

\bibitem{bai2025qwen2}
Bai, S., Chen, K., Liu, X., Wang, J., Ge, W., Song, S., Dang, K., Wang, P., Wang, S., Tang, J., et~al.: Qwen2. 5-vl technical report. arXiv preprint arXiv:2502.13923  (2025)

\bibitem{batra2025spatialthinker}
Batra, H., Tu, H., Chen, H., Lin, Y., Xie, C., Clark, R.: Spatialthinker: Reinforcing 3d reasoning in multimodal llms via spatial rewards. arXiv preprint arXiv:2511.07403  (2025)

\bibitem{cai2025spatialbot}
Cai, W., Ponomarenko, I., Yuan, J., Li, X., Yang, W., Dong, H., Zhao, B.: Spatialbot: Precise spatial understanding with vision language models. In: 2025 IEEE International Conference on Robotics and Automation (ICRA). pp. 9490--9498. IEEE (2025)

\bibitem{chandrasegaran2024hourvideo}
Chandrasegaran, K., Gupta, A., Hadzic, L.M., Kota, T., He, J., Eyzaguirre, C., Durante, Z., Li, M., Wu, J., Fei-Fei, L.: Hourvideo: 1-hour video-language understanding. Advances in Neural Information Processing Systems  \textbf{37},  53168--53197 (2024)

\bibitem{chen2024spatialvlm}
Chen, B., Xu, Z., Kirmani, S., Ichter, B., Sadigh, D., Guibas, L., Xia, F.: Spatialvlm: Endowing vision-language models with spatial reasoning capabilities. In: Proceedings of the IEEE/CVF Conference on Computer Vision and Pattern Recognition. pp. 14455--14465 (2024)

\bibitem{cheng2024spatialrgpt}
Cheng, A.C., Yin, H., Fu, Y., Guo, Q., Yang, R., Kautz, J., Wang, X., Liu, S.: Spatialrgpt: Grounded spatial reasoning in vision-language models. Advances in Neural Information Processing Systems  \textbf{37},  135062--135093 (2024)

\bibitem{chu2025sft}
Chu, T., Zhai, Y., Yang, J., Tong, S., Xie, S., Schuurmans, D., Le, Q.V., Levine, S., Ma, Y.: Sft memorizes, rl generalizes: A comparative study of foundation model post-training. arXiv preprint arXiv:2501.17161  (2025)

\bibitem{comanici2025gemini}
Comanici, G., Bieber, E., Schaekermann, M., Pasupat, I., Sachdeva, N., Dhillon, I., Blistein, M., Ram, O., Zhang, D., Rosen, E., et~al.: Gemini 2.5: Pushing the frontier with advanced reasoning, multimodality, long context, and next generation agentic capabilities. arXiv preprint arXiv:2507.06261  (2025)

\bibitem{daxberger2025mm}
Daxberger, E., Wenzel, N., Griffiths, D., Gang, H., Lazarow, J., Kohavi, G., Kang, K., Eichner, M., Yang, Y., Dehghan, A., et~al.: Mm-spatial: Exploring 3d spatial understanding in multimodal llms. In: Proceedings of the IEEE/CVF International Conference on Computer Vision. pp. 7395--7408 (2025)

\bibitem{deitke2025molmo}
Deitke, M., Clark, C., Lee, S., Tripathi, R., Yang, Y., Park, J.S., Salehi, M., Muennighoff, N., Lo, K., Soldaini, L., et~al.: Molmo and pixmo: Open weights and open data for state-of-the-art vision-language models. In: Proceedings of the Computer Vision and Pattern Recognition Conference. pp. 91--104 (2025)

\bibitem{driess2022learning}
Driess, D., Ha, J.S., Toussaint, M., Tedrake, R.: Learning models as functionals of signed-distance fields for manipulation planning. In: Conference on robot learning. pp. 245--255. PMLR (2022)

\bibitem{du2024embspatial}
Du, M., Wu, B., Li, Z., Huang, X.J., Wei, Z.: Embspatial-bench: Benchmarking spatial understanding for embodied tasks with large vision-language models. In: Proceedings of the 62nd Annual Meeting of the Association for Computational Linguistics (Volume 2: Short Papers). pp. 346--355 (2024)

\bibitem{fu2024blink}
Fu, X., Hu, Y., Li, B., Feng, Y., Wang, H., Lin, X., Roth, D., Smith, N.A., Ma, W.C., Krishna, R.: Blink: Multimodal large language models can see but not perceive. In: European Conference on Computer Vision. pp. 148--166. Springer (2024)

\bibitem{guo2025deepseek}
Guo, D., Yang, D., Zhang, H., Song, J., Zhang, R., Xu, R., Zhu, Q., Ma, S., Wang, P., Bi, X., et~al.: Deepseek-r1: Incentivizing reasoning capability in llms via reinforcement learning. arXiv preprint arXiv:2501.12948  (2025)

\bibitem{hong20233d}
Hong, Y., Zhen, H., Chen, P., Zheng, S., Du, Y., Chen, Z., Gan, C.: 3d-llm: Injecting the 3d world into large language models. Advances in Neural Information Processing Systems  \textbf{36},  20482--20494 (2023)

\bibitem{hu2022lora}
Hu, E.J., Shen, Y., Wallis, P., Allen-Zhu, Z., Li, Y., Wang, S., Wang, L., Chen, W., et~al.: Lora: Low-rank adaptation of large language models. Iclr  \textbf{1}(2), ~3 (2022)

\bibitem{huang2023embodied}
Huang, J., Yong, S., Ma, X., Linghu, X., Li, P., Wang, Y., Li, Q., Zhu, S.C., Jia, B., Huang, S.: An embodied generalist agent in 3d world. arXiv preprint arXiv:2311.12871  (2023)

\bibitem{huang20253d}
Huang, T., Zhang, Z., Tang, H.: 3d-r1: Enhancing reasoning in 3d vlms for unified scene understanding. arXiv preprint arXiv:2507.23478  (2025)

\bibitem{huang2024rekep}
Huang, W., Wang, C., Li, Y., Zhang, R., Fei-Fei, L.: Rekep: Spatio-temporal reasoning of relational keypoint constraints for robotic manipulation. arXiv preprint arXiv:2409.01652  (2024)

\bibitem{huang2025vision}
Huang, W., Jia, B., Zhai, Z., Cao, S., Ye, Z., Zhao, F., Xu, Z., Hu, Y., Lin, S.: Vision-r1: Incentivizing reasoning capability in multimodal large language models. arXiv preprint arXiv:2503.06749  (2025)

\bibitem{hurst2024gpt}
Hurst, A., Lerer, A., Goucher, A.P., Perelman, A., Ramesh, A., Clark, A., Ostrow, A., Welihinda, A., Hayes, A., Radford, A., et~al.: Gpt-4o system card. arXiv preprint arXiv:2410.21276  (2024)

\bibitem{kuhn1955hungarian}
Kuhn, H.W.: The hungarian method for the assignment problem. Naval research logistics quarterly  \textbf{2}(1-2),  83--97 (1955)

\bibitem{li2025viewspatial}
Li, D., Li, H., Wang, Z., Yan, Y., Zhang, H., Chen, S., Hou, G., Jiang, S., Zhang, W., Shen, Y., et~al.: Viewspatial-bench: Evaluating multi-perspective spatial localization in vision-language models. arXiv preprint arXiv:2505.21500  (2025)

\bibitem{li2025spatialladder}
Li, H., Li, D., Wang, Z., Yan, Y., Wu, H., Zhang, W., Shen, Y., Lu, W., Xiao, J., Zhuang, Y.: Spatialladder: Progressive training for spatial reasoning in vision-language models. arXiv preprint arXiv:2510.08531  (2025)

\bibitem{depthanything3}
Lin, H., Chen, S., Liew, J.H., Chen, D.Y., Li, Z., Shi, G., Feng, J., Kang, B.: Depth anything 3: Recovering the visual space from any views. arXiv preprint arXiv:2511.10647  (2025)

\bibitem{liu2023visual}
Liu, F., Emerson, G., Collier, N.: Visual spatial reasoning. Transactions of the Association for Computational Linguistics  \textbf{11},  635--651 (2023)

\bibitem{liu2023aerialvln}
Liu, S., Zhang, H., Qi, Y., Wang, P., Zhang, Y., Wu, Q.: Aerialvln: Vision-and-language navigation for uavs. In: Proceedings of the IEEE/CVF International Conference on Computer Vision. pp. 15384--15394 (2023)

\bibitem{liu2025ssr}
Liu, Y., Ma, M., Yu, X., Ding, P., Zhao, H., Sun, M., Huang, S., Wang, D.: Ssr: Enhancing depth perception in vision-language models via rationale-guided spatial reasoning. arXiv preprint arXiv:2505.12448  (2025)

\bibitem{liu2025spatial}
Liu, Y., Zhang, B., Zang, Y., Cao, Y., Xing, L., Dong, X., Duan, H., Lin, D., Wang, J.: Spatial-ssrl: Enhancing spatial understanding via self-supervised reinforcement learning. arXiv preprint arXiv:2510.27606  (2025)

\bibitem{liu2025nvila}
Liu, Z., Zhu, L., Shi, B., Zhang, Z., Lou, Y., Yang, S., Xi, H., Cao, S., Gu, Y., Li, D., et~al.: Nvila: Efficient frontier visual language models. In: Proceedings of the IEEE/CVF Conference on Computer Vision and Pattern Recognition. pp. 4122--4134 (2025)

\bibitem{liu2025visual}
Liu, Z., Sun, Z., Zang, Y., Dong, X., Cao, Y., Duan, H., Lin, D., Wang, J.: Visual-rft: Visual reinforcement fine-tuning. arXiv preprint arXiv:2503.01785  (2025)

\bibitem{ma2026thinking}
Ma, W., Sun, S., Yu, T., Wang, R., Chua, T.S., Bian, J.: Thinking with blueprints: Assisting vision-language models in spatial reasoning via structured object representation. arXiv preprint arXiv:2601.01984  (2026)

\bibitem{ma20253dsrbench}
Ma, W., Chen, H., Zhang, G., Chou, Y.C., Chen, J., de~Melo, C., Yuille, A.: 3dsrbench: A comprehensive 3d spatial reasoning benchmark. In: Proceedings of the IEEE/CVF International Conference on Computer Vision. pp. 6924--6934 (2025)

\bibitem{ma2025spatialreasoner}
Ma, W., Chou, Y.C., Liu, Q., Wang, X., Melo, C.M.d., Xie, J., Yuille, A.: Spatialreasoner: Towards explicit and generalizable 3d spatial reasoning. In: Advances in Neural Information Processing Systems. vol.~38 (2025)

\bibitem{ma2025spatialllm}
Ma, W., Ye, L., de~Melo, C.M., Yuille, A., Chen, J.: Spatialllm: A compound 3d-informed design towards spatially-intelligent large multimodal models. In: Proceedings of the Computer Vision and Pattern Recognition Conference. pp. 17249--17260 (2025)

\bibitem{meng2025mm}
Meng, F., Du, L., Liu, Z., Zhou, Z., Lu, Q., Fu, D., Han, T., Shi, B., Wang, W., He, J., et~al.: Mm-eureka: Exploring the frontiers of multimodal reasoning with rule-based reinforcement learning. arXiv preprint arXiv:2503.07365  (2025)

\bibitem{ouyang2025spacer}
Ouyang, K., Liu, Y., Wu, H., Liu, Y., Zhou, H., Zhou, J., Meng, F., Sun, X.: Spacer: Reinforcing mllms in video spatial reasoning. arXiv preprint arXiv:2504.01805  (2025)

\bibitem{shao2024deepseekmath}
Shao, Z., Wang, P., Zhu, Q., Xu, R., Song, J., Bi, X., Zhang, H., Zhang, M., Li, Y., Wu, Y., et~al.: Deepseekmath: Pushing the limits of mathematical reasoning in open language models. arXiv preprint arXiv:2402.03300  (2024)

\bibitem{shen2025vlm}
Shen, H., Liu, P., Li, J., Fang, C., Ma, Y., Liao, J., Shen, Q., Zhang, Z., Zhao, K., Zhang, Q., et~al.: Vlm-r1: A stable and generalizable r1-style large vision-language model. arXiv preprint arXiv:2504.07615  (2025)

\bibitem{song2025robospatial}
Song, C.H., Blukis, V., Tremblay, J., Tyree, S., Su, Y., Birchfield, S.: Robospatial: Teaching spatial understanding to 2d and 3d vision-language models for robotics. In: Proceedings of the Computer Vision and Pattern Recognition Conference. pp. 15768--15780 (2025)

\bibitem{tian2024drivevlm}
Tian, X., Gu, J., Li, B., Liu, Y., Wang, Y., Zhao, Z., Zhan, K., Jia, P., Lang, X., Zhao, H.: Drivevlm: The convergence of autonomous driving and large vision-language models. arXiv preprint arXiv:2402.12289  (2024)

\bibitem{tong2024cambrian}
Tong, P., Brown, E., Wu, P., Woo, S., IYER, A.J.V., Akula, S.C., Yang, S., Yang, J., Middepogu, M., Wang, Z., et~al.: Cambrian-1: A fully open, vision-centric exploration of multimodal llms. Advances in Neural Information Processing Systems  \textbf{37},  87310--87356 (2024)

\bibitem{wang2024omnidrive}
Wang, S., Yu, Z., Jiang, X., Lan, S., Shi, M., Chang, N., Kautz, J., Li, Y., Alvarez, J.M.: Omnidrive: A holistic llm-agent framework for autonomous driving with 3d perception, reasoning and planning. CoRR  (2024)

\bibitem{wang2025internvl3_5}
Wang, W., Gao, Z., Gu, L., Pu, H., Cui, L., Wei, X., Liu, Z., Jing, L., Ye, S., Shao, J., et~al.: Internvl3.5: Advancing open-source multimodal models in versatility, reasoning, and efficiency. arXiv preprint arXiv:2508.18265  (2025)

\bibitem{wu2025spatial}
Wu, D., Liu, F., Hung, Y.H., Duan, Y.: Spatial-mllm: Boosting mllm capabilities in visual-based spatial intelligence. arXiv preprint arXiv:2505.23747  (2025)

\bibitem{yang2025thinking}
Yang, J., Yang, S., Gupta, A.W., Han, R., Fei-Fei, L., Xie, S.: Thinking in space: How multimodal large language models see, remember, and recall spaces. In: Proceedings of the Computer Vision and Pattern Recognition Conference. pp. 10632--10643 (2025)

\bibitem{yang2025r1}
Yang, Y., He, X., Pan, H., Jiang, X., Deng, Y., Yang, X., Lu, H., Yin, D., Rao, F., Zhu, M., et~al.: R1-onevision: Advancing generalized multimodal reasoning through cross-modal formalization. arXiv preprint arXiv:2503.10615  (2025)

\bibitem{yang2024holodeck}
Yang, Y., Sun, F.Y., Weihs, L., VanderBilt, E., Herrasti, A., Han, W., Wu, J., Haber, N., Krishna, R., Liu, L., et~al.: Holodeck: Language guided generation of 3d embodied ai environments. In: Proceedings of the IEEE/CVF Conference on Computer Vision and Pattern Recognition. pp. 16227--16237 (2024)

\bibitem{yokoyama2024vlfm}
Yokoyama, N., Ha, S., Batra, D., Wang, J., Bucher, B.: Vlfm: Vision-language frontier maps for zero-shot semantic navigation. In: 2024 IEEE International Conference on Robotics and Automation (ICRA). pp. 42--48. IEEE (2024)

\bibitem{yu2025perception}
Yu, E., Lin, K., Zhao, L., Yin, J., Wei, Y., Peng, Y., Wei, H., Sun, J., Han, C., Ge, Z., et~al.: Perception-r1: Pioneering perception policy with reinforcement learning. arXiv preprint arXiv:2504.07954  (2025)

\bibitem{zhan2025actial}
Zhan, X., Huang, W., Sun, H., Fu, X., Ma, C., Cao, S., Jia, B., Lin, S., Yin, Z., Bai, L., et~al.: Actial: Activate spatial reasoning ability of multimodal large language models. arXiv preprint arXiv:2511.01618  (2025)

\bibitem{zhang2022focal}
Zhang, Y.F., Ren, W., Zhang, Z., Jia, Z., Wang, L., Tan, T.: Focal and efficient iou loss for accurate bounding box regression. Neurocomputing  \textbf{506},  146--157 (2022)

\bibitem{zheng2025multimodal}
Zheng, X., Dongfang, Z., Jiang, L., Zheng, B., Guo, Y., Zhang, Z., Albanese, G., Yang, R., Ma, M., Zhang, Z., et~al.: Multimodal spatial reasoning in the large model era: A survey and benchmarks. arXiv preprint arXiv:2510.25760  (2025)

\bibitem{zitkovich2023rt}
Zitkovich, B., Yu, T., Xu, S., Xu, P., Xiao, T., Xia, F., Wu, J., Wohlhart, P., Welker, S., Wahid, A., et~al.: Rt-2: Vision-language-action models transfer web knowledge to robotic control. In: Conference on Robot Learning. pp. 2165--2183. PMLR (2023)

\end{thebibliography}

\clearpage % 让附录从新的一页开始
\appendix

\section{Additional Dataset Construction Details}

\begin{table}[htbp]
    \centering
    \caption{Question templates for spatial reasoning tasks.}
    \label{tab:qa_template}
    \renewcommand{\arraystretch}{1.5}
    % 【修改点1】：在这里的 c 和 X 之间加上了 |
    \begin{tabularx}{\textwidth}{@{} c | X @{}}
    \toprule
    % 【修改点2】：将 c 改为 c@{}，保持右侧无多余边距
    \textbf{Task} & \multicolumn{1}{c@{}}{\textbf{Question Template}} \\
    \midrule
    % 注意这里把 [c] 改成了 [t]，实现顶部对齐
    \makecell[t]{Relative \\ Distance \\ Prediction} & 
    \textit{Which object is closer to the \textbf{\{reference object\}}, the \textbf{\{object a\}} or the \textbf{\{object b\}}? (Options: (A) \textbf{\{object a\}}, (B) \textbf{\{object b\}}, (C) they are equidistant)} \\
    \midrule
    % 这里同样改成了 [t]
    \makecell[t]{Perspective \\ Transformation \\ Reasoning} & 
    \textit{Assuming a camera is positioned at the \textbf{\{camera position\}} and facing \textbf{\{facing direction\}}, according to this camera's perspective, where is the \textbf{\{target object\}} located (\textbf{\{choice a\}}, \textbf{\{choice b\}}, \textbf{\{choice c\}}, or \textbf{\{choice d\}})?} \\
    \bottomrule
    \end{tabularx}
\end{table}

\subsection{Structured Chain-of-Thought Generation}
Structured Chain-of-Thought (CoT) spatial reasoning question-answer pairs are synthesized utilizing the raw data sourced from EmbSpatial and STVQA. These generated question-answer pairs span from fundamental spatial relations to complex spatial reasoning challenges. \Cref{tab:qa_template} details the question templates utilized to generate the structured CoT data. The generation process involves the following tasks:

\begin{enumerate}
    \item \textbf{Spatial Relation Understanding}: This task relies on the existing answer annotations from the EmbSpatial dataset, thereby eliminating the necessity to regenerate corresponding questions and answers. We start by generating appropriate captions and scene background information. Subsequently, the reasoning process is directly synthesized based on the 3D coordinates and refined.
    
    \item \textbf{Relative Distance Prediction}: This task involves a target object alongside two candidate objects. We establish a distance threshold of 0.03 meters. Variations below this threshold are classified as equidistant scenarios, whereas differences exceeding it definitively identify the strictly closer object as the ground truth. Based on their 3D spatial coordinates, we compute the Euclidean distance from the target to each candidate to determine the nearest one. 
    
    \item \textbf{Perspective Transformation Reasoning}: This task places a virtual camera at a reference object with a designated orientation. The questions are systematically categorized by 3D spatial offsets: significant horizontal displacement with negligible depth difference yields lateral queries (e.g., left/right); predominant depth variance leads to longitudinal queries (e.g., front/back); and substantial offsets across both axes produce composite directional queries (e.g., front-left or back-right). We compute the relative coordinates based on the x-axis and z-axis, then apply a perspective transformation to construct the reasoning process from the camera's point of view. 
    
    \item \textbf{Object-Centric Spatial Reasoning}: Similar to the first task, the data are extracted from the STVQA answer annotations. The initial stage of the processing pipeline is utilized to extract scene information and captions, effectively bypassing the question-answer generation step.
    
\end{enumerate}

\subsection{Data Refinement}

A comprehensive curation pipeline is deployed to maximize the reliability of the proposed dataset. During question answer generation, we formulate exactly one specific question type for each image in the EmbSpatial training set, which can actively preventing environmental overfitting. For the caption phase, we prompt Qwen-VL-MAX to construct detailed caption of scene. The development of reasoning trajectories is initially grounded in five predefined templates across eleven question types. To avoid rigid linguistic patterns and foster generalized reasoning capabilities, large language model is introduced to systematically rephrase and diversify these initial paths. To improve question answer quality, we employ Qwen-VL-MAX to generate answer based on our scene information, and strictly retain instances where the generated answers match the ground truth labels. Finally, human experts verify the remaining samples to confirm the logical consistency between the reasoning steps and the final conclusions. Following QA quality assessments, a subset of the data generated by Qwen-VL-MAX is selected and employed as supplementary data for cold-start SFT. The prompt we used are provided in the subsequent box.

\begin{promptbox}{Caption Generation Prompt}
\textbf{System Prompt:} \textbf{Task Context:} \\
- Question: \texttt{\{prompt\_text\}} \\[0.5em]
\textbf{Instructions:} \\
Output the response in the following structure. Briefly introduce the scene in natural language in the \texttt{<caption> </caption>} tags.
\end{promptbox}

\begin{promptbox}{Reasoning Refinement Prompt}
\textbf{[System Prompt]} \\[0.5em]
You are an expert in Chain-of-Thought reasoning optimization. Your task is to rewrite the `Draft Analysis' to make it natural, coherent, and logical, while STRICTLY PRESERVING all facts, numbers, equations, and object names.

\hrule % 绘制一条水平分隔线，用来区分 System 和 Use

\textbf{[User Prompt]} \\[0.5em]
\textbf{Context:} \\
Question: \texttt{\{question\}} \\
Ground Truth Answer: \texttt{\{answer\}} \\[0.8em]
\textbf{Draft Analysis (To be rewritten):} \\
\texttt{\{draft\_analysis\}} \\[0.8em]
\textbf{Instruction:} \\
Rewrite the Draft Analysis above. Combine the calculation steps and the logical deduction into a fluent paragraph. Do NOT change any coordinate values or calculation results. Make it sound like a step-by-step reasoning process.
\end{promptbox}

\begin{promptbox}{Answer Generation Prompt}
\textbf{System Prompt:} \textbf{Task Context:} \\
- Question: \texttt{\{prompt\_text\}} \\[0.5em]
\textbf{Metadata (Ground Truth):} \\
- Scene Objects: \texttt{\{scene\_info\_str\}} \\[0.5em]
\textbf{Precision:} BBox = \textbf{Integer}; Depth = \textbf{2 decimal places}. \\[0.5em]
\textbf{Instructions:} \\
Output the response in the following structure. Do NOT mention you were provided the ground truth or meta data.

Briefly introduce the scene in natural language in the \texttt{<caption> </caption>} tags and output the perception result in the specific JSON format: \texttt{<scene>[\{"label": "red apple", "bbox\_2d": [250, 400, 350, 550], "depth": 2.0\}, ...]</scene>}. Output JSON lists in a compact, single-line format (no newlines inside JSON).

Then, reason through the scene information to generate reasoning process in the \texttt{<analyze> </analyze>} tags. Output the final option in \texttt{<answer> </answer>} tags.
\end{promptbox}

\subsection{The Statics of SCOUT-24k}

\begin{table}[htbp]
    \centering
    \caption{Detailed statistics of our proposed SCOUT-24k dataset.}
    \label{tab:dataset_stats}
    \renewcommand{\arraystretch}{1.3} % 稍微增加行高，让表格更舒展
    \begin{tabular}{@{} l c c c c c @{}} % @{} 去除表格左右两端的额外空白；c 让数字与表头对齐
        \toprule
        \textbf{Subset} & \makecell[c]{\textbf{Relation} \\ \textbf{Understanding}} & \makecell[c]{\textbf{Relative} \\ \textbf{Distance}} & \makecell[c]{\textbf{Perspective} \\ \textbf{Transformation}} & \makecell[c]{\textbf{Object} \\ \textbf{Centric}} & \textbf{Total} \\
        \midrule
        SFT & 2,111 & 1,500 & 1,441 & 1,000 & 6,052 \\
        RL  & 6,917 & 3,403 & 6,352 & 1,942 & 18,614 \\
        \bottomrule
    \end{tabular}
\end{table}

The distribution of spatial reasoning tasks in our constructed SCOUT-24k is presented in \cref{tab:dataset_stats}.

\section{Additional Implement Details}

\subsection{Prompt Configuration for SCOUT}
The specific user prompt designed for the SCOUT framework is detailed in the text box below. To maintain strict alignment between the training and evaluation phases, the specific prompt is applied uniformly across Supervised Fine-Tuning (SFT), Reinforcement Learning (RL), and final inference.

\begin{promptbox}{SCOUT Template}
\textbf{User Prompt:} \textit{Image: \textless image\textgreater \\
Image size: (\textbf{\{Width\}} $\times$ \textbf{\{Height\}})\\
Question: \textbf{\{question\}} \\
Please output your thinking process wrapped in \textless think\textgreater \dots \textless /think\textgreater\ tags before the final answer. Briefly describe the image in \textless caption\textgreater\dots\textless /caption\textgreater\ tags and output detected objects with their bounding boxes and depth values in \textless scene\textgreater \dots \textless /scene\textgreater\ tags. Then analyze the question logically in \textless analyze\textgreater \dots \textless /analyze\textgreater\ tags to derive the answer and output the final answer in \textless answer\textgreater \dots \textless /answer\textgreater\ tags.}
\end{promptbox}

\begin{table}[t]
    \caption{Hyperparameter used in SFT training.}
    \label{tab:sft_hyperparameter}
  \centering
  \begin{tabular}{ll}
    \toprule
    \textbf{Hyperparameter} & \textbf{Value} \\
    \midrule
    lora\_rank & 8 \\
    cutoff\_length & 16384 \\
    per\_device\_train\_batch\_size & 1 \\
    gradient\_accumulation\_steps & 16 \\
    bf16 & true \\
    gradient\_checkpointing & true \\
    learning\_rate & $1\times 10^{-4}$ \\
    lr\_scheduler\_type & cosine \\
    warmup\_ratio & 0.1 \\
    num\_train\_epochs & 1 \\
    \bottomrule
  \end{tabular}
\end{table}

\begin{table}[t]
  \centering
  \caption{Hyperparameter used in RL training.}
  \label{tab:rl_hyperparameters}
  \begin{tabular}{ll}
    \toprule
    \textbf{Hyperparameter} & \textbf{Value} \\
    \midrule
    num\_generations & 8 \\
    temperature & 1.0 \\
    learning\_rate & $1\times 10^{-6}$ \\
    per\_device\_train\_batch\_size & 1 \\
    bf16 & true \\
    global\_batch\_size & 128 \\
    num\_train\_steps & 200 \\
    $\beta$ & 0.01 \\
    \bottomrule
  \end{tabular}
\end{table}

\subsection{Details of SFT and Reinforcement Learning}  

All training experiments are conducted on a hardware cluster comprising 7 $\times$ NVIDIA L20 GPUs (48GB memory per device). The training of SCOUT is divided into two sequential stages: a SFT cold-start and a subsequent reinforcement learning phase.

\subsubsection{SFT Cold-Start Initialization}

To endow the model with fundamental instruction following capabilities and the ability to produce structured outputs, we conduct a preliminary SFT phase. This cold-start procedure guarantees that the model can generate coherent, task-relevant responses, which serves as a critical prerequisite for the subsequent RL optimization. Implementation is carried out using the LLaMA-Factory framework, with the specific hyperparameters detailed in \cref{tab:sft_hyperparameter}.

\subsubsection{Reinforcement Learning Phase}

Building upon the initial SFT model, we apply our proposed RL methodology. Crucially, we perform a full-parameter update during this stage. This advanced training phase is executed via the EasyR1 framework, and all corresponding hyperparameters are specified in \cref{tab:rl_hyperparameters}.

\section{Additional Evaluation Details}
\begin{table*}[t]
\centering
\caption{Detailed Accuracy Results on 3DSRBench (\%)}
\label{tab:3dsrbench_accuracy}
\renewcommand{\arraystretch}{1.1}
\begin{tabular}{l|cccc|c}
\toprule
\multirow{2}{*}{\textbf{Model}} & \multicolumn{4}{c|}{\textbf{3DSRBench Tasks}} & \multirow{2}{*}{\textbf{Avg.}} \\
 & Height & Location & Orientation & Multi-Object & \\
\midrule
\rowcolor{rowgray} \multicolumn{6}{c}{\textit{Proprietary Models}} \\
GPT-4o & 54.9 & 59.8 & 21.8 & 39.5 & 44.78 \\
\hline
\rowcolor{rowgray} \multicolumn{6}{c}{\textit{Open Source and Specialized Spatial Models}} \\
Intern-VL3.5-4B & 37.43 & 51.89 & 38.1 & 36.57 & 42.1 \\
Intern-VL3.5-8B & 44.29 & 49.26 & 35.24 & 38.06 & 42.06 \\
SpaceLLaVA & 41.71 & 52 & 23.43 & 43.31 & 42.02 \\
SpatialBot & 50.29 & 60.69 & 20.19 & 44.57 & 45.83 \\
\hline
\rowcolor{rowgray} \multicolumn{6}{c}{\textit{Spatial Models Based on Qwen2.5-VL-3B}} \\
Qwen2.5-VL-3B & 38 & 52.69 & 33.9 & 33.71 & 40.65 \\
SpaceOm-3B & 36.29 & 54.29 & 32 & 35.2 & 41.07 \\
SpatialLadder-3B & 43.71 & 52.46 & 25.33 & 42.17 & 42.44 \\
SpatialThinker-3B & 37.71 & 56.11 & 33.9 & 38.4 & 43.31 \\
\rowcolor{rowgreen} SCOUT-3B (Ours) & 42.86 & 55.31 & 32 & 40.57 & 44.08 \\
\hline
\rowcolor{rowgray} \multicolumn{6}{c}{\textit{Spatial Models Based on Qwen2.5-VL-7B}} \\
Qwen2.5-VL-7B & 46.57 & 59.54 & 36.57 & 38.29 & 46.13 \\
SpatialReasoner-7B & 46 & 68.91 & 47.43 & 43.31 & 53.03 \\
SpatialThinker-7B & 44.29 & 57.83 & 37.9 & 41.49 & 46.59 \\
\rowcolor{rowgreen} SCOUT-7B (Ours) & 41.43 & 63.66 & 34.29 & 43.89 & 48.23 \\
\bottomrule
\end{tabular}
\end{table*} 

\subsection{Benchmarks.}
To comprehensively evaluate the spatial reasoning capabilities of the proposed model, we select six distinct benchmarks spanning from foundational spatial tasks to complex reasoning scenarios. For general spatial understanding, Embspatial, CV-Bench, and BLINK are employed. Specifically, Embspatial assesses basic spatial relationships (e.g., left/right, above/below, far/close); CV-Bench measures 2D spatial relations, object counting, depth ordering, and distance reasoning; and the Relative Depth subset of BLINK focuses on depth perception. To evaluate 3D capabilities, 3DSRBench is specifically adopted for spatial reasoning in real world. Finally, to probe spatial reasoning within embodied environments, we utilize SpatialBench and RoboSpatial. SpatialBench broadly covers spatial comprehension across object existence, positional relationships, physical interactions (e.g., reachability), and size comparisons. Furthermore, evaluations on the Configuration and Compatibility subsets of RoboSpatial simulate complex embodied tasks that demand a holistic understanding of object-to-object relationships.

\subsection{Baselines} 

To evaluate the proposed SCOUT models, we compare them against a diverse set of baselines grouped into three categories:
(1) \textbf{Proprietary Models}: We include GPT-4o, which represents the current state-of-the-art in commercial multimodal reasoning. It serves as an upper bound for spatial generalization under closed-source training regimes.
(2) \textbf{Open-Source General VLMs}: We evaluate Qwen2.5-VL (3B and 7B), representing cutting-edge architectures that offer strong general visual reasoning but lack specific spatial tuning. Additionally, we include InternVL-3.5 (4B and 8B), an advanced VLM featuring enhanced training strategies and superior data quality compared to its predecessors.
(3) \textbf{Specialized Spatial VLMs}: We compare with several open-source models explicitly tailored for spatial reasoning. These include SpaceLLaVA (a public re-implementation of SpatialVLM) and SpatialBot, which integrates RGB-D inputs for robust spatial perception. Furthermore, to enable a direct comparison within the Qwen2.5-VL ecosystem, we evaluate four Qwen-based spatial models: SpatialThinker (fine-tuned for structured spatial reasoning), SpaceOm (incorporating deeper chain-of-thought traces and Robo2VLM data), SpatialLadder (employing a progressive framework to enhance spatial perception), and SpatialReasoner (optimized via reinforcement learning and explicit 3D representations).

\subsection{Detailed Experiments Result}

The detailed spatial reasoning results on 3DSRBench, ViewSpatial, and VSI-Bench are provided in \cref{tab:3dsrbench_accuracy,tab:viewspatial_accuracy,tab:vsi_bench_detailed}.

\begin{table*}[t]
\centering
\caption{Detailed Accuracy Results on ViewSpatial (\%)}
\label{tab:viewspatial_accuracy}
% 如果表格超出页面宽度，\resizebox 会自动将其缩放到页面宽度以内
\resizebox{\textwidth}{!}{
\begin{tabular}{l ccc cccc}
\toprule
\multirow{2}{*}{\textbf{Models}} & \multicolumn{3}{c}{\textbf{Camera perspective}} & \multicolumn{4}{c}{\textbf{Person perspective}} \\
\cmidrule(lr){2-4} \cmidrule(lr){5-8}
& \makecell{Relative\\Direction} & \makecell{Object View\\Orientation} & \textbf{Total} & \makecell{Scene Simulation\\Relative Direction} & \makecell{Object View\\Orientation} & \makecell{Relative\\Direction} & \textbf{Total} \\
\midrule
Qwen-2.5-VL-3B & 44.22 & 31.93 & 39.80 & 26.70 & 41.16 & 31.00 & 32.82 \\
SCOUT-3B  & 46.87 & 29.02 & 40.45 & 28.24 & 49.50 & 35.39 & 37.48 \\
Qwen-2.5-VL-7B & 47.66 & 30.32 & 41.42 & 27.87 & 41.57 & 35.99 & 34.83 \\
SCOUT-7B  & 50.14 & 22.59 & 40.23 & 26.87 & 55.62 & 41.33 & 40.74 \\
\bottomrule
\end{tabular}
}
\end{table*}

\begin{table*}[t]
\centering
\caption{Detailed Accuracy Results on VSI-BENCH (\%)}
\label{tab:vsi_bench_detailed}
% 使用 resizebox 自动缩放以适应页面宽度
\resizebox{\textwidth}{!}{
\begin{tabular}{l ccccc cccccccc}
\toprule
\multirow{2}{*}{\textbf{Models}} & \multicolumn{5}{c}{\textbf{NQ}} & \multicolumn{7}{c}{\textbf{MCQ}} \\
\cmidrule(lr){2-6} \cmidrule(lr){7-13}
& \makecell{obj\\count} & \makecell{obj size\\est} & \makecell{room size\\est} & \makecell{obj abs\\dist} & \textbf{AVG} 
& \makecell{rel dir\\(hard)} & \makecell{rel dir\\(medium)} & \makecell{rel dir\\(easy)} & \makecell{rel\\dist} & \makecell{appear\\order} & \makecell{route\\planning} & \textbf{AVG} \\
\midrule
Qwen-2.5-VL-3B & 15.24 & 25.83 & 31.91 & 28.38 & 25.03 & 30.56 & 30.95 & 32.26 & 31.97 & 32.20 & 31.44 & 31.65 \\
SCOUT-3B  & 44.53 & 11.98 & 13.19 & 12.54 & 19.26 & 26.27 & 32.54 & 47.47 & 31.69 & 33.82 & 32.47 & 32.97 \\
Qwen-2.5-VL-7B & 47.20 & 48.30 & 45.45 & 25.41 & 40.52 & 23.86 & 26.98 & 47.47 & 41.55 & 35.76 & 30.93 & 34.94 \\
SCOUT-7B  & 59.15 & 18.24 & 27.60 & 22.47 & 29.35 & 30.56 & 47.35 & 51.15 & 39.85 & 32.68 & 30.41 & 38.07 \\
\bottomrule
\end{tabular}
}
\end{table*}

\section{Limitations and Future Work}

While the proposed approach demonstrates promising results, it remains subject to several limitations. First, due to computational resource constraints, our experiments are currently restricted to models at the 3B and 7B parameter scales. Second, the datasets rely on annotations of bounding boxes and object labels, lacking broader modalities such as multi-image contexts or video data. Finally, the proposed reinforcement learning method requires a strictly structured chain-of-thought format to extract perception results, which restricts flexibility and may be sub-optimal for other visual reasoning tasks.

To address these limitations, future work can focus on several key directions. First, scaling the framework to larger foundation models (e.g., 14B, 32B, and beyond) will facilitate the investigation of how increased model capacity affects spatial reasoning and RL optimization. Second, integrate multi-image and video modalities will extend our research from static 2D understanding to dynamic, spatiotemporal reasoning and incorporate locations and captions generated by VLMs will reduce the reliance on dense bounding box and label annotations. Finally, relaxing the strict dependency on a structured CoT format may enable more flexible, free-form reasoning pathways that generalize across a broader spectrum of complex visual reasoning tasks.

\end{document}